%% file: robustness_neurips_arxiv.tex
\documentclass{article}

\PassOptionsToPackage{numbers, sort&compress}{natbib}

     \usepackage[final]{neurips_2021}

\usepackage[utf8]{inputenc} 
\usepackage[T1]{fontenc}    
\usepackage{hyperref}       
\usepackage{url}            
\usepackage{booktabs}       
\usepackage{amsfonts}       
\usepackage{nicefrac}       
\usepackage{microtype}      
\usepackage{xcolor}         

\usepackage{graphicx}
\usepackage{subcaption}
\usepackage{booktabs} 
\usepackage{multirow}
\usepackage{tabularx}
\usepackage{colortbl}
\input{math_commands.tex}

\usepackage{hyperref}

\makeatletter
\newcommand*{\textoverline}[1]{$\overline{\hbox{#1}}\m@th$}
\makeatother
\newcommand{\newC}{\textoverline{C}}

\usepackage{xcolor}

\newcolumntype{x}[1]{>{\centering\arraybackslash}p{#1pt}}
\newcolumntype{y}[1]{>{\raggedright\arraybackslash}p{#1pt}}
\newcolumntype{z}[1]{>{\raggedleft\arraybackslash}p{#1pt}}
\newlength\savewidth\newcommand\shline{\noalign{\global\savewidth\arrayrulewidth
  \global\arrayrulewidth 1pt}\hline\noalign{\global\arrayrulewidth\savewidth}}

\newcommand{\err}[2]{\small #1\!{\fontsize{6}{1} \selectfont $\scriptscriptstyle \pm$#2}}
\newcommand{\phantomerr}[2]{\small #1\!{\fontsize{6}{1} \selectfont \phantom{$\scriptscriptstyle \pm$#2}}}
\definecolor{Gray}{gray}{0.92}
\newcolumntype{g}[1]{>{\columncolor{Gray}\centering\arraybackslash}p{#1pt}}

\title{On Interaction Between Augmentations and Corruptions in Natural Corruption Robustness}

\author{%
  Eric Mintun\thanks{This work completed as part of the Facebook AI residency program.} \\
  Facebook AI Research\\
  \texttt{mintun@fb.com} \\
\And
  Alexander Kirillov \\
  Facebook AI Research\\
  \texttt{akirillov@fb.com} \\
\And
  Saining Xie \\
  Facebook AI Research\\
  \texttt{s9xie@fb.com} \\
}

\begin{document}

\maketitle

\begin{abstract}
Invariance to a broad array of image corruptions, such as warping, noise, or color shifts, is an important aspect of building robust models in computer vision. 
Recently, several new data augmentations have been proposed that significantly improve performance on ImageNet-C, a benchmark of such corruptions. 
However, there is still a lack of basic understanding on the relationship between data augmentations and test-time corruptions. 
To this end, we develop a feature space for image transforms, and then use a new measure in this space between augmentations and corruptions called the Minimal Sample Distance to demonstrate a strong correlation between similarity and performance.
We then investigate recent data augmentations and observe a significant degradation in corruption robustness when the test-time corruptions are sampled to be perceptually dissimilar from ImageNet-C in this feature space.
Our results suggest that test error can be improved by training on perceptually similar augmentations, and data augmentations may not generalize well beyond the existing benchmark. 
We hope our results and tools will allow for more robust progress towards improving robustness to image corruptions. 
We provide code at \url{https://github.com/facebookresearch/augmentation-corruption}.
\end{abstract}

\section{Introduction}

Robustness to distribution shift, \emph{i.e.} when the train and test distributions differ, is an important feature of practical machine learning models. Among many forms of distribution shift, one particularly relevant
category for computer vision are image corruptions. For example, test data may come from sources that differ from the training set in terms of lighting, camera quality, or other features.  Post-processing transforms, such as photo touch-up, image filters, or compression effects are commonplace in real-world data. Models developed using clean, undistorted inputs typically perform dramatically worse when confronted with these sorts of image corruptions~\cite{hendrycks2018benchmarking, geirhos2018generalisation}. The subject of corruption robustness has a long history in computer vision~\cite{simard1998transformation,bruna2013invariant,dodge2017study} and recently has been studied actively with the release of benchmark datasets such as ImageNet-C \cite{hendrycks2018benchmarking}.  

One particular property of image corruptions is that they are low-level distortions in nature. Corruptions are transformations of an image that affect structural information such as colors, textures, or geometry~\cite{ding2020image} and are typically free of high-level semantics. Therefore, it is natural to expect that \emph{data augmentation} techniques, which expand the training set with random low-level transformations, can help learn robust models. Indeed, data augmentation has become a central technique in several recent methods \cite{hendrycks2019augmix,lopes2019improving, rusak2020increasing} that achieve large improvements on ImageNet-C and related benchmarks.

\begin{figure*}
    \centering
    \includegraphics[scale=1]{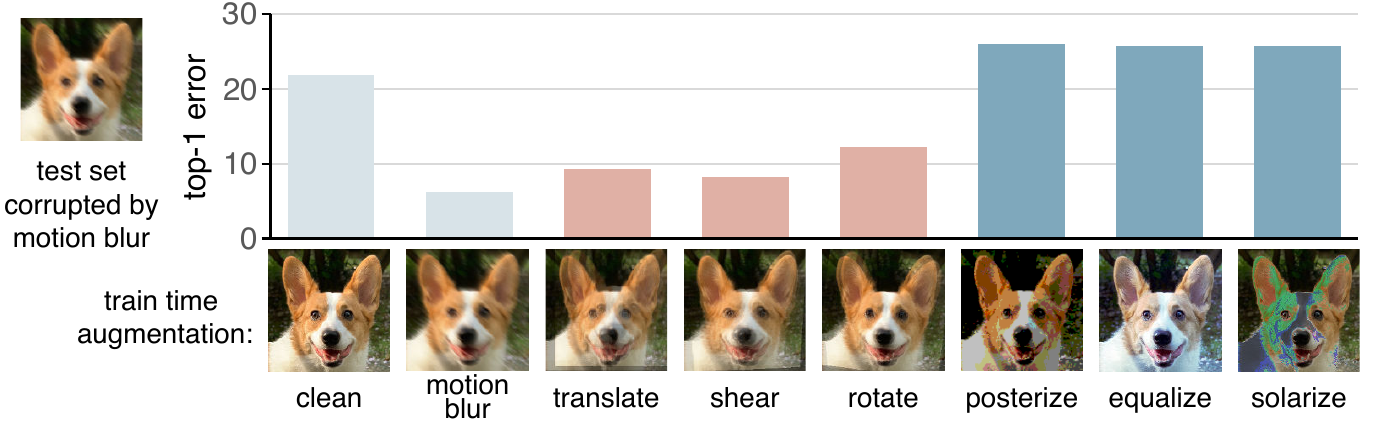}
    \caption{
        \textbf{A toy experiment.} We train multiple models on CIFAR-10 \cite{krizhevsky2009learning} using different augmentation schemes. Each scheme is based on a single basic image transformation type and enhanced by overlaying random instantiations of the transformation for each input image following \citet{hendrycks2019augmix}. We compare these models on the CIFAR-10 test set corrupted by the motion blur, a corruption used in the ImageNet-C corruption benchmark~\cite{hendrycks2018benchmarking}. None of the augmentation schemes contains motion blur; however, the models trained with geometric-based augmentations significantly outperform the baseline model trained on the clean images while color-based augmentations show no gains. We note the geometric augmentations can produce a result visually similar to a blur by overlaying copies of shifted images\protect\footnotemark.}
    \label{fig:teaser}
    \vskip -0.1in
\end{figure*}

One caveat for data augmentation based approaches is the test corruptions are expected to be \emph{unknown} at training time. If the corruptions are known, they may simply be applied to the training set as data augmentations to trivially adapt to the test distribution. Instead, an ideal robust model needs to be robust to \emph{any} valid corruption, including ones unseen in any previous benchmark. Of course, in practice the robustness of a model can only be evaluated approximately by measuring its corruption error on a representative corruption benchmark.  To avoid trivial adaptation to the benchmark, recent works manually exclude test corruptions from the training augmentations. However, with a toy experiment presented in Figure~\ref{fig:teaser}, we argue that this strategy alone might not be enough and that visually similar augmentation outputs and test corruptions can lead to significant benchmark improvements even if the exact corruption transformations are excluded.

This observation raises two important questions. One, \emph{how exactly does the similarity between train time augmentations and corruptions of the test set affect the error?} And two, if the gains are due to the similarity, they may not translate into better robustness to other possible corruptions, so \emph{how well will data augmentations generalize beyond a given benchmark?} In this work, we take a step towards answering these questions, with the goal of better understanding the relationship between data augmentation and test-time corruptions. Using a feature space on image transforms and a new measure called Minimal Sample Distance (MSD) on this space, we are able to quantify the distance between augmentation schemes and classes of corruption transformation. With our approach, we empirically show an intuitive yet surprisingly overlooked finding:
\newline\newline
\emph{Augmentation-corruption perceptual similarity is a strong predictor of corruption error.}\footnotetext{Example transforms are for illustrative purpose only and are exaggerated.  Base image \copyright~Sehee Park.}
\newline\newline
Based on this finding, we perform additional experiments to show that data augmentation aids corruption robustness by increasing perceptual similarity between a (possibly small) fraction of the training data and the test set. To further support our claims, we introduce a set of new corruptions, called CIFAR/ImageNet-\newC{}, to test the degree to which common data augmentation methods generalize from the original CIFAR/ImageNet-C.  To choose these corruptions, we expand the set of natural corruptions and sample new corruptions that are far away from CIFAR/ImageNet-C in our feature space for measuring perceptual similarity.  We then demonstrate that augmentation schemes designed specifically to improve robustness show significantly degraded performance on CIFAR/ImageNet-\newC{}.  Some augmentation schemes still show some improvement over baseline, which suggests meaningful progress towards general corruption robustness is being made, but different augmentation schemes exhibit different degrees of generalization capability.  As an implication, caution is needed for fair robustness evaluations when additional data augmentation is introduced.

These results suggest a major challenge that is often overlooked in the study of corruption robustness: \emph{generalization is often poor}. Since perceptual similarity can predict performance, for any fixed finite set of test corruptions, improvements on that set may generalize poorly to dissimilar corruptions.  We hope that these results, tools, and benchmarks will help researchers better understand \emph{why} a given augmentation scheme has good corruption error and whether it should be expected to generalize to dissimilar corruptions.  On the positive side, our experiments show that \emph{generalization does emerge} among perceptually similar transforms, and that only a \emph{small fraction} of sampled augmentations need to be similar to a given corruption. Section \ref{sec:discussion} discusses these points in more depth.

\section{Related Work}
\label{sec:relatedwork}
\paragraph{Corruption robustness benchmarks and analysis.}  ImageNet-C \cite{hendrycks2018benchmarking} is a corruption dataset often used as a benchmark in robustness studies.  Other corruption datasets \cite{hendrycks2020many, shankar2019image} collect corrupted images from real world sources and thus have a mixture of semantic distribution shifts and perceptual transforms.  Corruption robustness differs from adversarial robustness \cite{szegedy2013intriguing}, which seeks invariance to small, worst case distortions.  One notable difference is that improving corruption robustness often slightly improves regular test error, instead of harming it.  \citet{yin2019fourier} analyzes corruption robustness in the context of transforms' frequency spectra; this can also influence corruption error independently from perceptual similarity.  Here we study the relationship between augmentations and corruptions more generically, and explore the relationship between perceptual similarity and generalization to new corruptions. \citet{dao2019kernel} and \citet{wu2020generalization} study the theory of data augmentation for regular test error. \citet{hendrycks2020many} and \citet{taori2020measuring} study how the performance on synthetic corruption transforms generalizes to performance on corruption datasets collected from the real world.  Here we do not address this issue directly but touch upon it in the discussion.
\paragraph{Improving corruption robustness.}  Data augmentations designed to improve robustness include AugMix \cite{hendrycks2019augmix}, which composites common image transforms, Patch Gaussian \cite{lopes2019improving}, which applies Gaussian noise in square patches, and ANT \cite{rusak2020increasing}, which augments with an adversarially learned noise distribution.  AutoAugment \cite{cubuk2019autoaugment} learns augmentation policies that optimize clean error but has since been shown to improve corruption error \cite{yin2019fourier}.  Mixup \cite{zhang2018mixup} can improve robustness \cite{lee2020compounding}, but its label augmentation complicates the dependence on image augmentation.  Stylized-ImageNet \cite{geirhos2018imagenet}, which applies style transfer to input images, can also improve robustness.  DeepAugment \cite{hendrycks2020many}, which applies augmentations to a deep representation of an image, can also give large improvements in robustness. Noisy Student \cite{xie2020self} and Assemble-ResNet \cite{lee2020compounding} combine data augmentation with new models and training procedures and greatly enhance corruption robustness. In addition to training-time methods, there are approaches that adapt to unseen corruptions at test time, e.g. using self-supervised tasks \cite{sun2020test}, entropy minimization \cite{wang2020tent}, or with a focus on privacy and data transmission efficiency \cite{liang2020we}. While we do not directly address these approaches here, our methods potentially provide tools that could be used to measure shifting distributions in an online regime. 
\section{Perceptual similarity for augmentations and corruptions}
\label{sec:transformsimilarity}
First, we study the importance of similarity between augmentations and corruptions for improving performance on those corruptions.  To do so, we need a means to compare augmentations and corruptions.  Both types of transforms are perceptual in nature, meaning they affect low-level image structure while leaving high-level semantic information intact, so we expect a good distance to be a measure of \emph{perceptual similarity}.  Then, we need to find the appropriate measure of distance between the augmentation and corruption \emph{distributions}.  We will argue below that distributional equivalence is not appropriate in the context of corruption robustness, and instead introduce the \emph{minimal sample distance}, a simple measure that does capture a relevant sense of distribution distance.
\paragraph{Measuring similarity between perceptual transforms.}  We define a perceptual transform as a transform that acts on low-level image structure but not high-level semantic information.  As such, we expect two transforms should be similar if their actions on this low-level structure are similar, independent of algorithmic or per-pixel differences between them.  A closely related, well-studied problem is the perceptual similarity between \emph{images}.  A common approach is to train a neural network on a classification task and use intermediate layers as a feature space for measuring distances \cite{zhang2018unreasonable}.  We adapt this idea to obtain a feature space for measuring distances between perceptual transforms.

We start with a feature extractor for images, which we call $\hat{f}(x)$.  To train the model from which we will extract features, we assume access to a dataset $\mathbb{D}$ of image label pairs $(x,y)$ associated with a classification task.  The model should be trained using only default data augmentation for the task in question so that the feature extractor is independent of the transforms we will use it to study.  In order to obtain a very simple measure, we use just the last hidden layer of the network as a feature space.

A perceptual transform $t(x)$ may be encoded by applying it to all images in $\sD$, encoding the transformed images, and averaging the features over these images.  For efficiency, we find it sufficient to average over only a randomly sampled subset of images $\sD_S$ in $\sD$.  In Section \ref{subsec:exsetup} we discuss the size of $\sD_S$.  The random choice of images is a property of the feature extractor, and so remains fixed when encoding multiple transforms.  This reduces variance when computing distances between two transforms.  The transform feature extractor is given by $f(t) = \mathbb{E}_{x \in \sD_S}  [ \hat{f} (t(x)) - \hat{f}(x) ]$.  The \emph{perceptual similarity} between an augmentation and a corruption can be taken as the $L_2$ distance on this feature space $f$. 
\paragraph{Minimal sample distance.} We now seek to compare the distribution of an augmentation scheme $p_a$ to a distribution of a corruption benchmark $p_c$.  If the goal was to optimize error on a \emph{known} corruption distribution, exact equivalence of distributions is the correct measure to minimize.  But since the goal is robustness to general, \emph{unknown} corruption distributions, a good augmentation scheme should be equivalent to no single corruption distribution.

\begin{figure}
    \centering
    \includegraphics[scale=1.0]{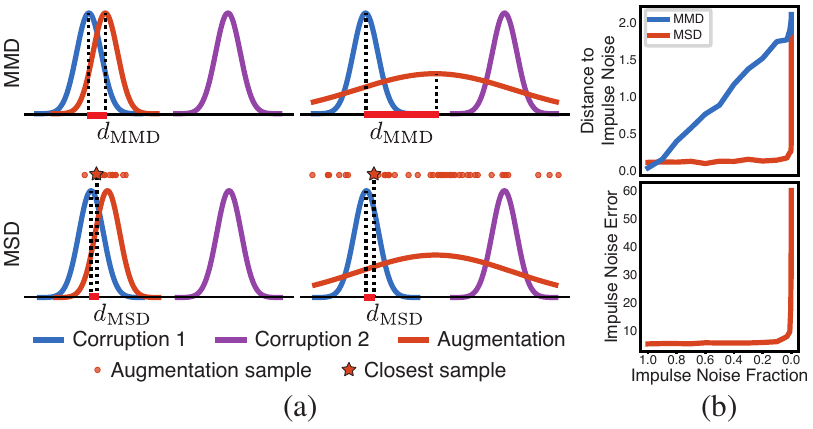}
    \vskip -0.1in
    \caption{(a) Schematic comparison of MMD to MSD.  MMD measures the distance between distribution centers and is only small if the augmentation overlaps with a corruption.  MSD measures to the nearest sampled point in the set of samples (marked by a star) and is small even for broad distributions that overlap with multiple corruptions.  (b) We test on images corrupted with \emph{impulse noise}, and train on images augmented with a mixture of \emph{impulse noise} and \emph{motion blur}.  As the mixing fraction of \emph{impulse noise} decreases, MMD between the augmentation and corruption grows linearly while MSD and error stay low until nearly 0\% mixing fraction.}
    \label{fig:mmdteaser}
    \vskip -0.1in
\end{figure}

To illustrate this behavior, consider a toy problem where we have access to the corruption transforms at training time.  A very rough, necessary-but-insufficient measure of distributional similarity is $d_{\mathrm{MMD}}(p_a,p_c) = || \E_{a\sim p_a} [ f(a) ]  - \E_{c\sim p_c} [ f(c) ] ||$.  This is the maximal mean discrepancy on a fixed, finite feature space, so for brevity we will refer to it as MMD.  We still employ the featurization $f(t)$, since we are comparing transforms and not images, unlike in typical domain adaptation.  Consider two corruption distributions, here \emph{impulse noise} and \emph{motion blur}, and an augmentation scheme that is a mixture of the two corruption distributions.  Figure \ref{fig:mmdteaser}b shows MMD between the augmentation and \emph{impulse noise} corruption scales linearly with mixing fraction, but error on \emph{impulse noise} remains low until the mixing fraction is almost 0\% impulse noise.  This implies distributional similarity is a poor predictor of corruption error.  Indeed, low $d_{\mathrm{MMD}}$ with any one corruption distribution suggests the augmentation overlaps it significantly, so the augmentation is unlikely to aid dissimilar corruptions.

Our expectation for the behavior of the error in Figure \ref{fig:mmdteaser}b is that networks can often successfully memorize rare examples seen during training, so that only a very small fraction of sampled images need \emph{impulse noise} augmentations to perform well on \emph{impulse noise} corruptions.  An appropriate distance should then measure how close augmentation samples can come to the corruption distribution, even if the density of those samples is low.  We thus propose a very simple measure called \emph{minimal sample distance (MSD)}, which is just the perceptual similarity between an average corruption and the closest augmentation from a finite set of samples $\sA \sim p_a$:
\begin{equation}
    d_{\mathrm{MSD}}(p_a, p_c) = \min_{a \in \sA \sim p_a} || f(a)  - \E_{c \sim p_c} [ f(c) ] || \, .
\end{equation}
A schematic comparison of MMD and MSD is shown in Figure \ref{fig:mmdteaser}a.  While both MMD and MSD are small for an augmentation scheme that is distributionally similar to a corruption distribution, only MSD remains small for a broad distribution that occasionally produces samples near multiple corruption distributions. Figure \ref{fig:mmdteaser}b shows MSD, like test error, is small for most mixing fractions in the toy problem described above.  Note the measure's need to accommodate robustness to general, unknown corruption distributions has led it to be asymmetric, so it differs from more formal distance metrics that may be used to predict generalization error, such as the Wasserstein distance \cite{zilly2019frechet}.

 \section{Perceptual similarity is predictive of corruption error}
 \label{sec:simpredictserr}
We are now equipped to measure how important this augmentation-corruption similarity is for corruption error.  For a large number of augmentation schemes, we will measure both the MSD to a corruption distribution and the corruption error of a model trained with that scheme.  We will find a correlation between MSD and corruption error, which provides evidence that networks generalize across perceptually similar transforms.  Then, we will calculate MSD for augmentation schemes in the literature that have been shown to improve error on corruption benchmarks.  We will find a correlation between MSD and error here as well, suggesting their success is in part explained by their perceptual similarity to the benchmark.  This implies there may be a risk of poor generalization to different benchmarks, since we would not expect this improvement to transfer to a dissimilar corruption.

\subsection{Experimental setup}
\label{subsec:exsetup}
\paragraph{Corruptions.} We use CIFAR-10-C \cite{hendrycks2018benchmarking}, which is a common benchmark used for studying corruption robustness.  It consists of 15 corruptions, each further split into five different severities of transformation, applied to the CIFAR-10 test set.  The 15 corruptions fall into four categories: per-pixel noise, blurring, synthetic weather effects, and digital transforms.  We treat each corruption at each severity as a separate distribution for the sake of calculating MSD and error; however, for simplicity we average errors and distances over severity to present a single result per corruption.

\paragraph{Space of augmentation schemes.} To build each sampled augmentation transform, we will composite a set of base augmentations.  For base augmentations, we consider the nine common image transforms used in \citet{hendrycks2019augmix}.  There are five geometric transforms and four color transforms.  By taking all subsets of these base augmentations, we obtain $2^9=512$ unique augmentation schemes, collectively called the \emph{augmentation powerset}.  Also following \citet{hendrycks2019augmix}, we composite transforms in two ways:  by applying one after another, or by applying them to copies of the image and then linearly superimposing the results. Examples of both augmentations and corruptions are provided in Appendix \ref{app:transforms}.
\begin{figure}
    \centering
    \includegraphics[scale=1.0]{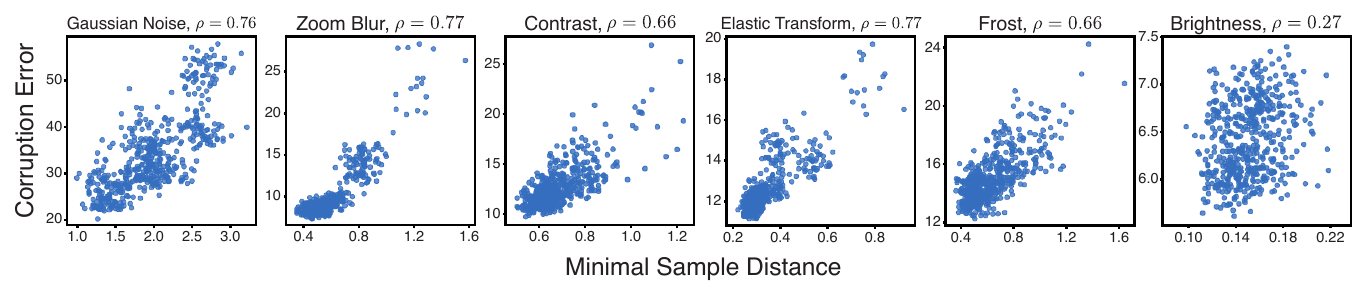}
    \vskip -0.1in
    \caption{Example relationships between MSD and corruption error.  $\rho$ is the Spearman rank correlation.  MSD correlates well with error across all four categories of corruption in CIFAR-10-C. For completeness, we also show \emph{brightness}, a negative example where correlation is poor.}
    \vskip -0.1in
    \label{fig:msd_only}
\end{figure}
\paragraph{Computing similarity and corruption error.} A WideResNet-40-2 \cite{BMVC2016_87} model is pre-trained on CIFAR-10 using default augmentation and training parameters from \citet{hendrycks2019augmix}.  WideResNet is a common baseline model used when studying data augmentation on CIFAR-10 \cite{hendrycks2019augmix, cubuk2019autoaugment, zhang2018mixup}.  Its last hidden layer is used as the feature space.  For MSD, we average over 100 images, 100 corruptions, and minimize over 100k augmentations. With this number of corruptions and images, we find that the average standard deviation in distance between an augmentation and the averaged corruptions is roughly five percent of the mean, which is smaller than the typical feature in our results found below, given in Figure \ref{fig:msd_only}. We also find that using VGG \cite{simonyan2014very} instead of WideResNet for the feature extractor gives similar results. Details for these calculations are in Appendix \ref{app:ablation}. Images for calculating MSD are from the training set and do not have default training augmentation. A WideResNet-40-2 with the same training parameters is used for corruption error evaluation.

\subsection{Analysis}
\paragraph{MSD correlates with corruption error.} First, we establish the correlation between MSD and corruption error on the augmentation powerset.  MSD shows strong correlation with corruption error across corruptions types in all four categories of CIFAR-10-C, and for a large majority of CIFAR-10-C corruptions in general: 12 of 15 have Spearman rank correlation greater than 0.6. Figure \ref{fig:msd_only} shows the relationship between distance and corruption error on six example corruptions, including one negative example for which correlation is low. A complete set of plots is below in Figure \ref{fig:distverrorreal}.  This corruption, \emph{brightness}, may give poor results because it is a single low-level image statistic that can vary significantly from image to image, and thus may not be well represented by our feature extractor. Appendix \ref{app:msdexp} has a few supplemental experiments. First, we we confirm MMD correlates poorly with corruption error, as expected. In particular, we expect broad augmentation schemes produce samples similar to a larger set of corruptions, leading to both lower MSD and lower corruption error but higher MMD. Second, we repeat our experiment but do not train on the augmentations, instead only adapting the batch norm statistics of a pre-trained model to them. We still find a strong correlation, suggesting our methods are compatible with the results of \citet{schneider2020improving}, which shows such an adaptation of the batch norm statistics to a corruption can improve corruption error.
\begin{figure}
    \centering
    \includegraphics[scale=1.0]{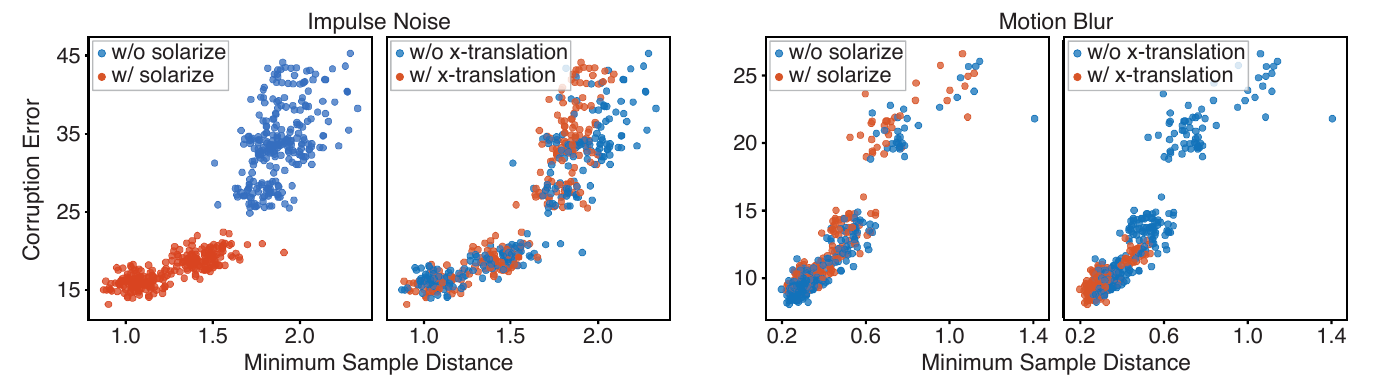}
    \vskip -0.1in
    \caption{Example relationships between base augmentations and corruptions.  Including \emph{solarize} reduces MSD on the perceptually similar \emph{impulse noise} corruption. Including \emph{x translation} reduces MSD on the perceptually similar \emph{motion blur} corruption.  MSD is not decreased for dissimilar augmentation-corruption pairs.} 
    \label{fig:distverrorcolored}
    \vskip -0.1in
\end{figure}
\paragraph{An example of perceptual similarity.} Here we illustrate the perceptual nature of the similarity measure, using an example with two base augmentations and two corruptions.  The augmentation \emph{solarize} and the corruption \emph{impulse noise} both insert bright pixels into the image, though in different ways.  Linear superpositions of the augmentation \emph{x translation} are visually similar to a blur, such as the corruption \emph{motion blur}.  Figure \ref{fig:distverrorcolored} shows MSD vs error where augmentation schemes that include \emph{solarize} and \emph{x translation} are colored.  It is clear that including an augmentation greatly decreases MSD to its perceptually similar corruption, while having little effect on MSD to its perceptually dissimilar corruption.

\paragraph{MSD and corruption error in real augmentation methods.} The augmentation powerset may be used as a baseline for comparing real data augmentation schemes.  Figure \ref{fig:distverrorreal} shows MSD-error correlations for Patch Gaussian \cite{lopes2019improving}, AutoAugment \cite{cubuk2019autoaugment}, and Augmix \cite{hendrycks2019augmix}, along with the cloud of augmentation powerset points for all 15 CIFAR-10-C corruptions.  The real augmentation schemes follow the same general trend that lower error predicts lower MSD.  A few intuitive correlations are also captured in Figure \ref{fig:distverrorreal}. Patch Gaussian has low MSD to noise corruptions. AutoAugment, which contains contrast and Gaussian blurring augmentations in its sub-policies, has low MSD with \emph{contrast} and \emph{defocus blur}.  A negative example is \emph{fog}, on which MSD to AutoAugment is not predictive of corruption error.

This correlation suggests generalization may be poor beyond an existing benchmark, since an augmentation scheme may be perceptually similar to one benchmark but not another. For augmentations and corruptions that are explicitly the same, such as $\emph{contrast}$ in AutoAugment and ImageNet-C, this is typically accounted for by removing such transforms from the augmentation scheme when testing corruption robustness\footnote{For this analysis, we wish to treat explicit transform similarity and perceptual transform similarity on the same footing, so we choose not to remove these overlapping augmentations.}. But in addition to these explicit similarities, Figure \ref{fig:distverrorreal} shows quantitatively that perceptual similarity between non-identical augmentations and corruptions is also strongly predictive of corruption error. This includes possibly unexpected similarities, such as between Patch Gaussian and \emph{glass blur}, which introduces random pixel-level permutations as noise. This suggests that perceptually similar augmentations and corruptions should be treated with the same care as identical transforms. In particular, tools such as MSD help us determine \emph{why} an augmentation scheme improves corruption error, so we can better understand if new methods will generalize beyond their tested benchmarks. Next we test this generalization by finding corruptions dissimilar to ImageNet-C.

\section{ImageNet-\newC{}: benchmarking with dissimilar corruptions}
\label{sec:newdataset}
We now introduce a set of corruptions, called ImageNet-\newC{}, that are perceptually dissimilar to ImageNet-C in our transform feature space, and we will show that several augmentation schemes have degraded performance on the new dataset. We emphasize that the dataset selection method uses only default data augmentation and was fixed before we looked at the results for different augmentations, so we are not adversarially selecting against the tested augmentation schemes.
\begin{figure}
    \centering
    \includegraphics[scale=1.0]{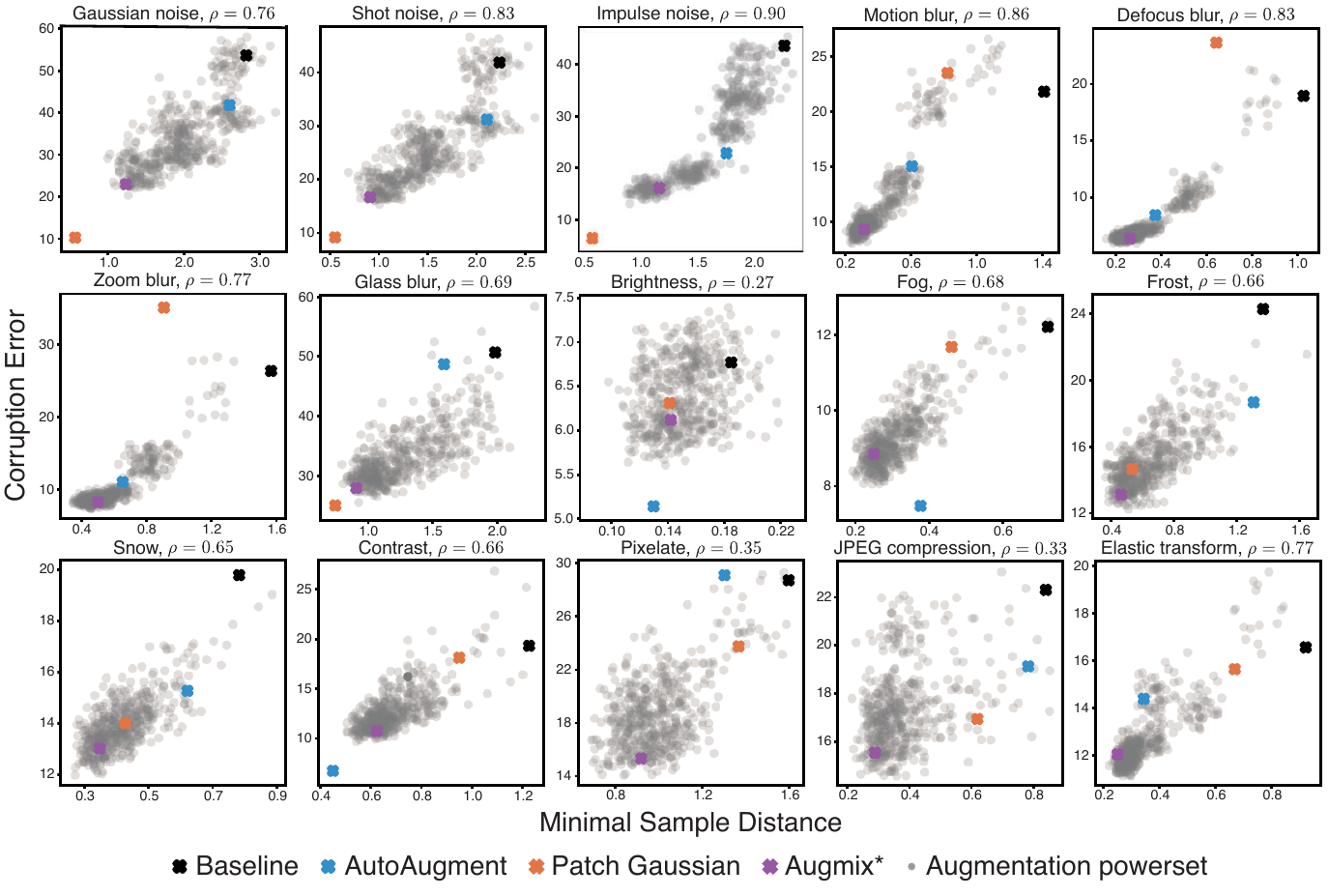}
    \vskip -0.1in
    \caption{Correlations for augmentation schemes from the literature.  Patch Gaussian is similar to noise, while AutoAugment is similar to contrast and blur, as expected from their formulation.  Glass blur acts more like a noise corruption than a blur for these augmentation schemes, likely because it randomly permutes pixels. As a negative example, MSD does not correlate well with error for AutoAugment on \emph{fog}. \textsuperscript{*}AugMix here refers to just the augmentation distribution in \citet{hendrycks2019augmix}, not the proposed Jensen-Shannon divergence loss.}
    \label{fig:distverrorreal}
    \vskip -0.1in
\end{figure}
\paragraph{Dataset construction.} Here we present an overview of the dataset construction method. We build 30 new corruptions in 10 severities, from which the 10 most dissimilar corruptions will be chosen.  We adapt common filters and noise distributions available online \cite{jhlabs, filterpedia} to produce human interpretable images.  The transforms include warps, blurs, color distortions, noise additions, and obscuring effects.  Examples of the new corruptions and exact details of the construction method are provided in Appendices \ref{app:cbardetails} and \ref{app:transforms}.

\begin{figure}
    \centering
    \includegraphics[scale=1.0]{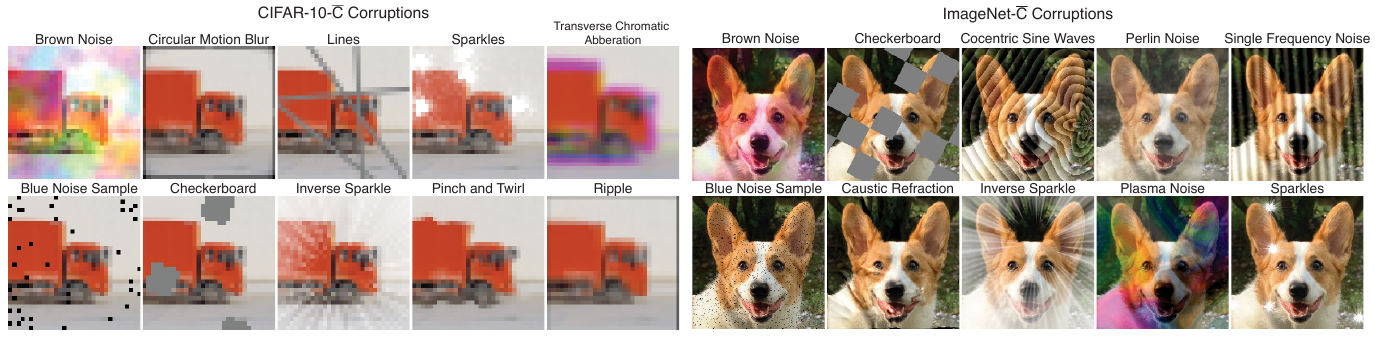}
    \vskip -0.1in
    \caption{Example CIFAR-10-\newC{} and ImageNet-\newC{} corruptions.  While still human interpretable, new corruptions are sampled to be dissimilar from CIFAR-10/ImageNet-C.  Base images \copyright~Sehee Park and Chenxu Han.}
    \label{fig:newdsets}
    \vskip -0.1in
\end{figure}

To assure that the new dataset is no harder than ImageNet-C, we restrict the average corruption error of the new dataset to be similar to that of ImageNet-C for default augmentation.  We then generate many potential datasets and measure the average shift in distance to ImageNet-C that each corruption contributes.  Note that while MSD is a measure between augmentations and corruptions, here we are comparing corruptions to other corruptions and thus use MMD in our transform feature space. ImageNet-\newC{} then consists of the 10 corruptions types with the largest average shift in distance.  Like ImageNet-C, each has five different severities, with severities chosen so that the average error matches ImageNet-C for default augmentation.  Example transforms from ImageNet-\newC{} and CIFAR-10-\newC{} are shown in Figure \ref{fig:newdsets}. This procedure in our feature space produces corruptions intuitively dissimilar from ImageNet-C and CIFAR-10-C. 

\paragraph{Results.}We test AutoAugment \cite{cubuk2019autoaugment}, Patch Gaussian \cite{lopes2019improving}, AugMix \cite{hendrycks2019augmix}, ANT\textsuperscript{3x3} \cite{rusak2020increasing}, Stylized-ImageNet \cite{geirhos2018imagenet}, and DeepAugment \cite{hendrycks2020many} on our new datasets and show results in Table \ref{tab:newdatasetresults}. CIFAR-10 models are WideResNet-40-2 with training parameters from \citet{hendrycks2019augmix}. ImageNet \cite{deng2009imagenet} models are ResNet-50 \cite{he2016deep} with training parameters from \citet{goyal2017accurate}.  Stylized-ImageNet is trained jointly with ImageNet for half the epochs and starts from a model pre-trained on ImageNet, following \citet{geirhos2018imagenet}.  Models use default data augmentation as well as the augmentation being tested, except ImageNet color jittering is not used.  All corruptions are applied in-memory instead of loaded from a compressed file; this can affect results especially on high frequency corruptions.  

Since Section \ref{sec:simpredictserr} suggests several augmentation schemes are perceptually similar to ImageNet-C corruptions, we might expect these methods to have worse error on the new corruptions. Indeed, every augmentation scheme performs worse.  Different augmentation schemes also degrade by significantly different amounts, from $+0.7\%$ for AutoAugment to $+7.3\%$ for PatchGaussian, which changes their ranking by corruption error and leads to inconsistency of generalization. In Table \ref{tab:bigmodels}, we compare performance on several robust models\cite{mahajan2018exploring,orhan2019robustness,xie2020self,tan2019efficientnet,dosovitskiy2020image,touvron2020deit,zhang2020resnest} that are not primarily augmentation-based and see no similar pattern of degradation, further suggesting that augmentation-corruption dissimilarity is the cause of the higher error.

Errors of individual corruptions in ImageNet-\newC{} are also revealing. For all augmentation schemes, there is significant improvement on \emph{blue sample noise}\footnote{This corruption is conceptually similar with \emph{impulse noise} but also gives a large distance; this may be a failure mode of our measure, maybe since \emph{impulse noise} has bright pixels and \emph{blue noise sample} has dark pixels.} but little improvement on \emph{sparkles} or \emph{inverse sparkles}.  Only AutoAugment does well on \emph{checkerboard}, perhaps because only AutoAugment's geometric transforms produce empty space, similar to \emph{checkerboard}'s occluded regions. These examples suggest a slightly different benchmark could yield significantly different results. Indeed, for a hypothetical benchmark that excluded \emph{blue sample noise} and \emph{checkerboard}, AutoAugment and Patch Gaussian have 57.3\% and 57.2\% error respectively, little better than baseline of 57.4\%. AugMix fairs only a little better with 54.3\% error. Even DeepAugment+AugMix, which is in general a strong augmentation scheme, shows a big discrepancy in performance across different corruptions, improving \emph{single frequency noise} by 31\%, but \emph{inverse sparkles} by only 2.3\%. Generalization to dissimilar corruptions is thus both inconsistent and typically quite poor. Single benchmarks and aggregate corruption scores are likely not enough for careful evaluation of robustness to unknown corruptions, and it is important to study why proposed augmentations succeed to better understand how well they might generalize.

It may be surprising that Stylized-ImageNet also degrades, given that it is intuitively very different from every corruption. While our measure works for augmentations, it does not cover all possible methods that improve robustness, such as more complicated algorithms like Stylized-ImageNet. Stylized-ImageNet degradation may be due to other reasons. For instance, it primarily augments texture information and may help mostly with higher frequency corruptions, as can be seen by its improvement on \emph{single frequency noise} and \emph{cocentric sine waves}; ImageNet-\newC{} has fewer such corruptions than ImageNet-C. ImageNet-\newC{} is thus a useful tool for understanding the interaction between training procedure and corruption distribution, even beyond perceptual similarity.

Nevertheless, note that it is the intuitively broader augmentation schemes, such as AutoAugment, AugMix, Stylized-ImageNet, and DeepAugment that generalize better to ImageNet-\newC{}.  The importance of breadth has also been explored elsewhere\cite{yin2019fourier, hendrycks2020many}, but in the previous sections we have provided new quantitative evidence for \emph{why} this may be true: broad augmentation schemes may be perceptually similar to more types of corruptions, and thus more likely to be perceptually similar to a new corruption. Moreover, AugMix and DeepAugment still improve over baseline on ImageNet-\newC{}, so there is reason to be optimistic that robustness to unknown corruptions is an achievable goal, as long as evaluation is treated carefully.

\begin{table*}
    \caption{Test error for several data augmentation methods on CIFAR-10-\newC{} and ImageNet-10-\newC{}, for which every method performs worse than on ImageNet-C or CIFAR-10-C.  The increase in error differs significantly between different augmentation methods. Descriptions of the abbreviations and standard deviations for individual corruptions are in Appendix \ref{app:cbardetails}. `Baseline' refers to default augmentation only. Averages are over five runs for ImageNet and ten for CIFAR-10. \textsuperscript{*}ANT, DeepAugment(DA) and DeepAugment+AugMix (DA+AM) use the pre-trained model provided with the associated papers and have different training parameters.}
    \label{tab:newdatasetresults}
    \vskip 0.03in
    \centering
    \small
    \setlength\tabcolsep{1.5pt}
    \begin{tabular}[t]{ x{36}  x{30}  x{30} g{34}  x{22} x{22} x{22} x{22} x{22} x{22} x{22} x{22} x{22} x{22} }
        & \textbf{IN-C} & \multicolumn{2}{c }{\textbf{IN-\newC{}}} & \multicolumn{10}{c }{\textbf{ ImageNet-\newC{} Corruptions} }  \\
        Aug & Err & Err & $\Delta$IN-C & BSmpl & Plsm & Ckbd & CSin & SFrq & Brwn & Prln & Sprk & ISprk & Rfrac \\
        \shline
        Baseline &  \err{58.1}{0.4} & \err{57.7}{0.2} & -0.4 & 68.6 & 71.7 & 49.4 & 84.7 & 79.0 & 37.5 & 34.3 & 32.4 & 76.7 & 42.8 \\ \hline
        AA & \err{55.0}{0.2} & \err{55.7}{0.3} & +0.7 & 54.8 & 68.3 & 43.8 & 86.5 & 78.8 & 34.5 & 33.8 & 36.1 & 77.1 & 43.8 \\ 
        SIN & \err{52.4}{0.1} & \err{55.8}{0.3} & +3.4 & 54.7 & 69.8 & 52.8 & 79.6 & 69.2 & 37.8 & 35.3 & 37.0 & 77.3 & 44.1 \\ 
        AugMix & \err{49.2}{0.7} & \err{52.4}{0.2} & +3.2 & 43.2 & 72.2 & 46.1 & 76.3 & 67.4 & 38.8 & 32.4 & 32.3 & 76.4 & 39.2 \\ 
        PG & \err{49.3}{0.2} & \err{56.6}{0.4} & +7.3 & 60.3 & 74.1 & 48.5 & 82.1 & 76.7 & 38.9 & 34.6 & 32.1 & 76.5 & 42.1 \\ 
        ANT* & \phantomerr{48.8}{0.2} & \phantomerr{53.9}{0.2} &  +5.1 & 35.8 & 75.5 & 56.9 & 76.4 & 63.7 & 41.0 & 35.2 & 35.0 & 76.1 & 43.3 \\ 
        DA* & \phantomerr{46.6}{0.2} & \phantomerr{51.0}{0.2} &  +4.4 & 41.7 & 73.3 & 53.9 & 74.6 & 50.9 & 37.2 & 30.3 & 32.9 & 74.7 & 40.9 \\ 
         DA+AM* & \phantomerr{41.0}{0.2} & \phantomerr{48.3}{0.2} &  +7.3 & 34.9 & 67.9 & 49.8 & 69.7 & 48.0 & 35.2 & 30.6 & 32.9 & 74.3 & 39.8 \\ 
    \end{tabular}
    \\
    \begin{tabular}[t]{ x{36}  x{30}  x{30} g{34}  x{22} x{22} x{22} x{22} x{22} x{22} x{22} x{22} x{22} x{22} }
        & \textbf{C10-C} & \multicolumn{2}{c }{\textbf{C10-\newC{}}} & \multicolumn{10}{c }{\textbf{CIFAR-10-\newC{} Corruptions}}  \\ 
        Aug & Err & Err & $\Delta$C10-C & BSmpl & Brwn & Ckbd & CBlur & ISprk & Line & P\&T & Rppl & Sprk & TCA \\
        \shline
        Baseline & \err{27.0}{0.6} &  \err{27.1}{0.5} & +0.1 & 42.9 & 27.2 & 23.3 & 11.8 & 43.3 & 26.2 & 11.3 & 21.6 & 21.0 & 42.9 \\ \hline
        AA & \err{19.4}{0.2} & \err{21.0}{0.4} & +1.6 & 17.7 & 17.5 & 17.6 & 9.5 & 40.4 & 23.6 & 10.7 & 23.5 & 17.5 & 31.8   \\ 
        AugMix & \err{11.1}{0.2} & \err{16.0}{0.3} & +5.9 & 9.8 & 27.8 & 13.4 & 5.9 & 30.3 & 18.0 & 8.3 & 12.1 & 15.5 & 19.2 \\
        PG & \err{17.0}{0.3} & \err{23.8}{0.5} & +6.8 & 9.0 & 30.1 & 21.6 & 12.8 & 35.4 & 20.6 & 8.8 & 21.5 & 19.3 & 59.5  \\ 
    \end{tabular}
\end{table*}
\begin{table}
\caption{Comparison of errors on ImageNet-C and ImageNet-\newC{} for several robust models: WSL (weakly supervised ResNeXt-101-32x8d \cite{mahajan2018exploring,orhan2019robustness}), EN (EfficientNet-B0 \cite{tan2019efficientnet}), NS (Noisy Student EN-B0 \cite{xie2020self}), ViT-S (Transformer \cite{dosovitskiy2020image,touvron2020deit}), ResNeSt (ResNeSt-50d, \cite{zhang2020resnest}), using pre-trained models provided with the respective papers. These models do not rely primarily on data augmentation to be robust, and there is no consistent degradation on ImageNet-\newC{}. This is additional evidence that the worse performance in Table \ref{tab:newdatasetresults} does not occur because ImageNet-\newC{} is harder generally.}    
\label{tab:bigmodels}
\vskip 0.05in
\centering
\begin{tabular}{cccccc}
  & WSL & EN & NS & ViT-S & ResNeSt \\ \shline
IN-C Err & 38.1 & 55.7 & 52.1 & 44.5 & 44.4 \\
IN-\newC{} Err & 39.2 & 53.4 & 52.2 & 41.1 &  41.6
\end{tabular}
\end{table}
\section{Discussion}
\label{sec:discussion}

\paragraph{\emph{Societal Impact.}} Our method for finding dissimilar corruptions could in principle be used to adversarially attack computer vision systems, such as those in content moderation or self-driving cars. Moreover, our ultimate goal is to help improve robustness in computer vision, and such robust systems may be used in detrimentals ways, for example in autonomous weapons or surveillance. However, we expect better evaluation of robust models to have definite benefits as well. In the long run, such an understanding should help defend against adversarial attacks. Our tools could also be used to challenge purportedly robust systems that are actually dangerously unreliable, such as an autonomous driving system that is robust to common corruption benchmarks yet fails to be robust to a dissimilar but important corruption, e.g., maybe glare. For instance, is the model employing data augmentation that is perceptually similar to the corruptions being used to report good robustness? Is the set of validation corruptions sufficiently broad that we would expect reasonable generalization to an unseen corruption? If we generate a dissimilar set of corruptions using the procedure we develop here, does the model still perform well on the new corruptions? Quantitative ways to answer these questions may provide a means to verify the robust performance of a model before it encounters and potentially fails on a critical, previously unseen corruption.

\paragraph{\emph{Corruption robustness as a secondary learning task.}} We have provided evidence that data augmentation may not generalize well beyond a given corruption benchmark. To explore this further, consider an analogy to a regular learning problem.  We may think of corruption robustness in the presence of data augmentation as a sort of secondary task layered on the primary classification task: the set of data augmentations is the training set, the set of corruptions is the test set, and the goal is to achieve invariance of the underlying primary task.  In this language, the `datasets' involved are quite small: ImageNet-C has only 15 corruption types, and several augmentation schemes composite only around 10 basic transforms.  In this case, standard machine learning practice would dictate a training/validation/test set split; it is only the size and breadth of modern vision datasets that has allowed this to be neglected in certain cases recently.  But the effective dataset size of a corruption robustness problem is tiny, so having a held-out test set seems necessary.  To emphasize, this is not a test set of the underlying classification task, for which generalization has been studied by \citet{recht2018cifar,recht2019imagenet}.  Instead, it is a test set of corruption transforms themselves.  This means there would be validation/test split of dissimilar transformations, both applied to the ImageNet validation set\footnote{The validation set provided in \citet{hendrycks2018benchmarking} consists of perceptually similar transforms to ImageNet-C and would not be expected to work well for the validation discussed here.}. 

\paragraph{\emph{Real-world corruption robustness.}} Recently, \citet{hendrycks2020many} and \citet{taori2020measuring} study how performance on corruption transforms generalizes to real-world corruptions and come to conflicting conclusions.  Though we do not study real-world corruptions, we have proposed a mechanism that may explain the conflict: performance will generalize between transforms and real-world corruptions if they are perceptually similar, but will likely not if they are dissimilar.  Since \citet{hendrycks2020many} and \citet{taori2020measuring} draw on different real-world and synthetic corruptions, it may be that the perceptual similarity between datasets differs in the two analyses.  This also suggests a way to find additional corruption transforms that correlate with real-world corruptions: transforms should be sought that have maximal perceptual similarity with real-world corruptions.

\paragraph{\emph{Generalization does occur.}} We have encountered two features of data augmentation that may explain why it can be such a powerful tool for corruption robustness, despite the issues discussed above.  First, within a class of perceptually similar transforms, generalization does occur.  This means each simple data augmentation may confer robustness to many complicated corruptions, as long as they share perceptual similarity.  Second, dissimilar augmentations in an augmentation scheme often causes little to no loss in performance, as long as a similar augmentation is also present.  We briefly study this in Appendix \ref{app:numaugmentations} by demonstrating that adding many dissimilar augmentations increases error much less than adding a few similar augmentations decreases it. These two features suggest broad augmentation schemes with many dissimilar augmentations may confer robustness to a large class of unknown corruptions.  More generally, we think data augmentation is a promising direction of study for corruption robustness, as long as significant care is taken in evaluation.

\section*{Acknowledgements and Funding Disclosure}
Eric Mintun would like to thank Matthew Leavitt, Sho Yaida, and Achal Dave for discussions during the development of this work. Additionally, he would like to acknowledge the Facebook AI residency program for providing excellent training and support in AI research. The authors received no external funding and have no competing interests.

\bibliography{robustness}
\bibliographystyle{icml2021}

\newpage

\appendix

\section{Sampling similar augmentations more frequently gives minor performance improvements}
\label{app:numaugmentations}
Here we describe an alternative experiment that shows how the introduction of dissimilar augmentations affects corruption error.  For a broad data augmentation scheme that provides robustness to many dissimilar corruptions, each corruption may only have a similar augmentation sampled some small fraction of the time.  This small fraction of samples must be sufficient to yield good performance on each corruption to obtain robustness overall.  We expect that this should be the case, since neural networks are often good at memorizing rare examples.  Additionally, the toy problem in Figure \ref{fig:mmdteaser} suggests that a large fraction of sampled augmentations may be dissimilar without significant loss in corruption error.  Here we show the effect using a real augmentation scheme.

\begin{figure}
    \centering
    \includegraphics[scale=1.0]{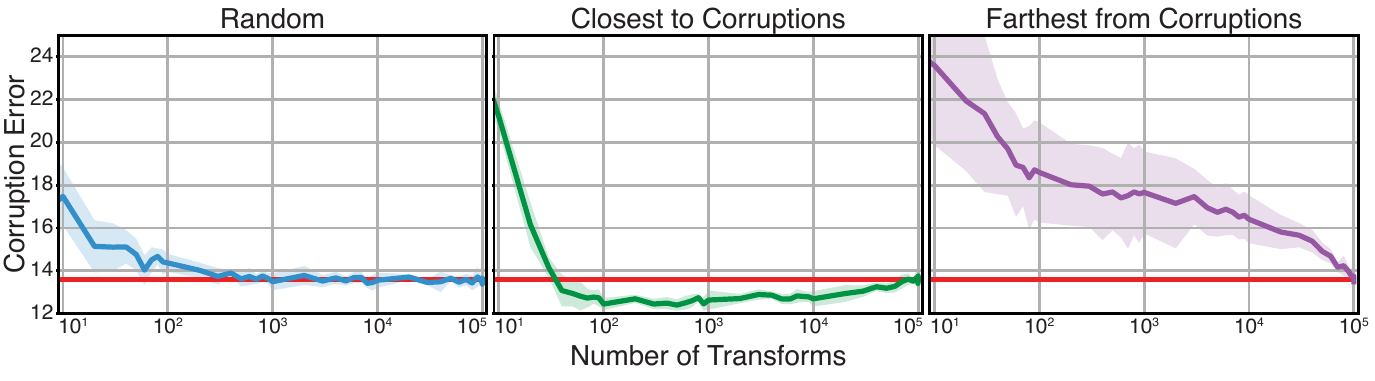}
    \caption{Average corruption error on ImageNet-C as a function the size of a fixed subset of AugMix augmentations.  During training, augmentations are only sampled from the subset.  The subset is chosen one of three ways: randomly, the most similar augmentations to ImageNet-C, or the least similar augmentations to ImageNet-C.  Choosing similar corruptions improves error beyond AugMix, but not by as much that choosing dissimilar augmentations harms it.}
    \label{fig:numaugmentations}
\end{figure}

We consider performance on CIFAR-10-C when training with AugMix augmentations (we do not use their Jensen-Shannon divergence loss, which gives additional improvements).  However, instead of sampling directly from the AugMix distribution during training, we first sample 100k transforms and sort these transforms by their distance to the CIFAR-10-C corruptions.  This sorting is done to evenly distribute the augmentations among the 75 (15 corruptions in 5 severities) individual corruptions; \emph{e.g.} the first 75 augmentations in the list are the closest augmentation to each corruption.  Then we take a fixed-size subset $\sA$ of these transforms and train on augmentations sampled only from this subset using the training parameters from \citet{hendrycks2019augmix}.  We select $\sA$ three different ways: randomly, taking the $|\sA|$ closest augmentations, and taking the $|\sA|$ farthest augmentations.  We then measure the average corruption error on CIFAR-10-C and plot this error against $|\sA|$ in Figure \ref{fig:numaugmentations}.

First, we note that for randomly sampled augmentations, $\sA$ does not need to be very large to match AugMix in performance.  Even though training on AugMix with our training parameters would normally would produce 5 million uniquely sampled augmentations, only around 1000 are needed to achieve equivalent performance.  Training on the closest augmentations exceeds regular AugMix performance with only around 100 unique transforms, which acts as additional evidence that augmentation-corruption similarity correlates with corruption error.  This gain in accuracy comes not from having access to better transformations, but from having more frequent access to them at training time.  However, the gain is fairly mild at only around 1\%, even though the best transformations are sampled all of the time instead of rarely.  The gain from frequency is much less than the gain from having more similar augmentations, since choosing the most dissimilar augmentations gives around a 5\% drop in accuracy.  This suggests that it is a net positive to decrease the frequency of sampling similar augmentations in order to include augmentations similar to another set of corruptions: the gain in accuracy on the new corruption set will likely out weight the small loss in accuracy on the original set.
\newpage
\section{Additional MSD and MMD experiments}
\label{app:msdexp}
\subsection{Comparison of MSD and MMD}
\begin{figure}
    \centering
    \includegraphics[scale=1.0]{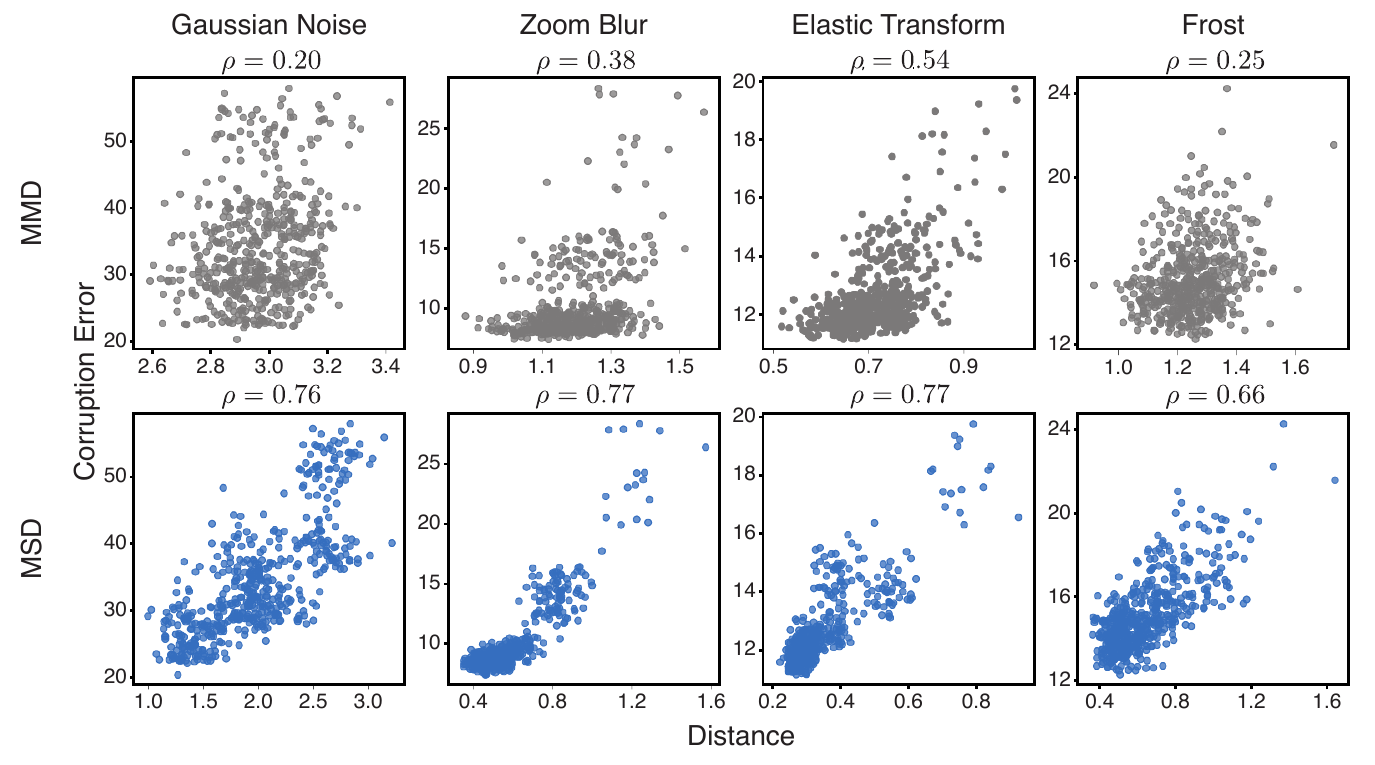}
    \caption{Example relationships between augmentation-corruption distance and corruption error for two distance scores, MMD and MSD.  $\rho$ is the Spearman rank correlation.  MMD between an augmentation and corruption distribution is not typically predictive of corruption error.  MSD correlates well across all four categories of corruption in CIFAR-10-C.}
    \label{fig:distverror}
\end{figure}

To support the use of MSD for comparing augmentations and corruptions, we confirm here that the more naive measure of MMD correlates poorly with corruption error. We calculate MMD and MSD as defined in Section \ref{sec:transformsimilarity} between each augmentation in the augmentation powerset and the corruptions in CIFAR-10-C. Figure \ref{fig:distverror} shows a comparison of how MMD and MSD correlate with corruption error on sample corruptions.  MMD typically shows poor correlation, while MSD has strong correlation in all four categories of corruption.
\subsection{Analyzing generalization with MMD}
\label{app:mmdoverfitting}
\begin{figure}
    \centering
    \includegraphics[scale=1.0]{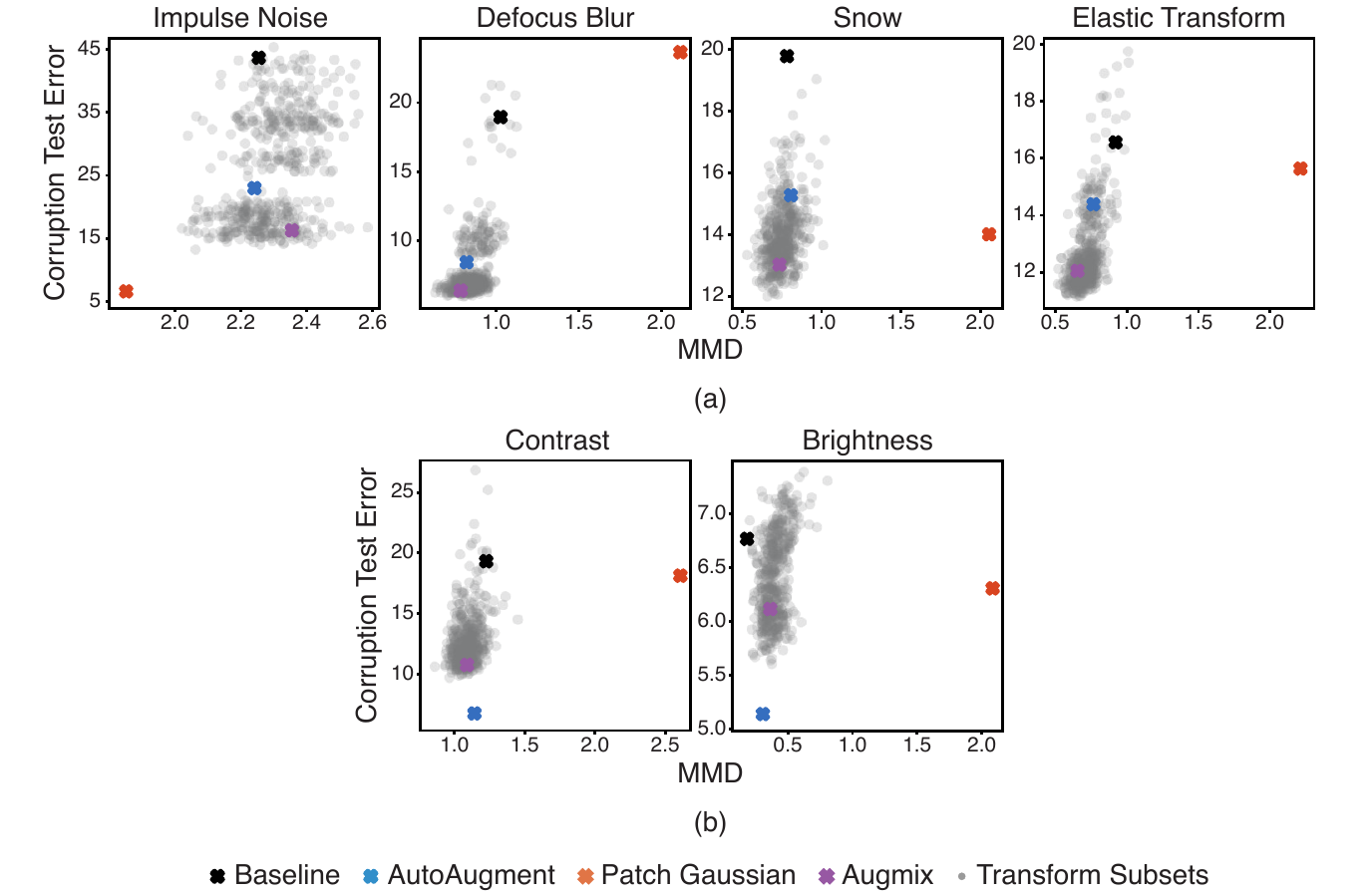}
    \caption{(a) Patch Gaussian shows a low MMD distance on the noise corruptions and a high MMD distance on every other corruption, suggesting that it may be perceptually similar to the noise corruptions in a distributional sense.  (b) While AutoAugment contains \emph{contrast} and \emph{brightness} augmentations, it is broad enough that it doesn't have a low MMD to these corruptions.  Note that since \emph{brightness} shows poor correlation for MSD, it is possible that in this case the MMD does not change for other reasons.}
    \label{fig:mmdverrorreal}
\end{figure}

In Section \ref{sec:transformsimilarity}, we argue distributional equivalence is usually not appropriate for studying augmentation-correlation similarity because augmentation distributions are typically broader than any one corruption distribution.  However, were an augmentation perceptually similar to a class of corruptions in the distributional sense, it might suggest at poor generalization to dissimilar corruptions. Using the simple, necessary but insufficient measure we call MMD in Section \ref{sec:transformsimilarity}, we can study a weak sense of distributional equivalence. Figure \ref{fig:mmdverrorreal} shows example MMD-error correlations.  For Patch Guassian, MMD is low for the noise corruptions and high for everything else, while AutoAugment and AugMix, which are constructed out of many visually distinct transforms, show no strong correlation.  This suggests the intuitive result that Patch Gaussian does not just have perceptual overlap with the noise corruptions, but is perceptually similar to them in a more distributional sense. We might then expect poorer generalization from Patch Gaussian to corruptions dissimilar from the noise corruptions, which includes ImageNet-C.

\subsection{MSD vs Batch-Norm Adaptation}
\begin{figure}
    \centering
    \includegraphics[scale=1]{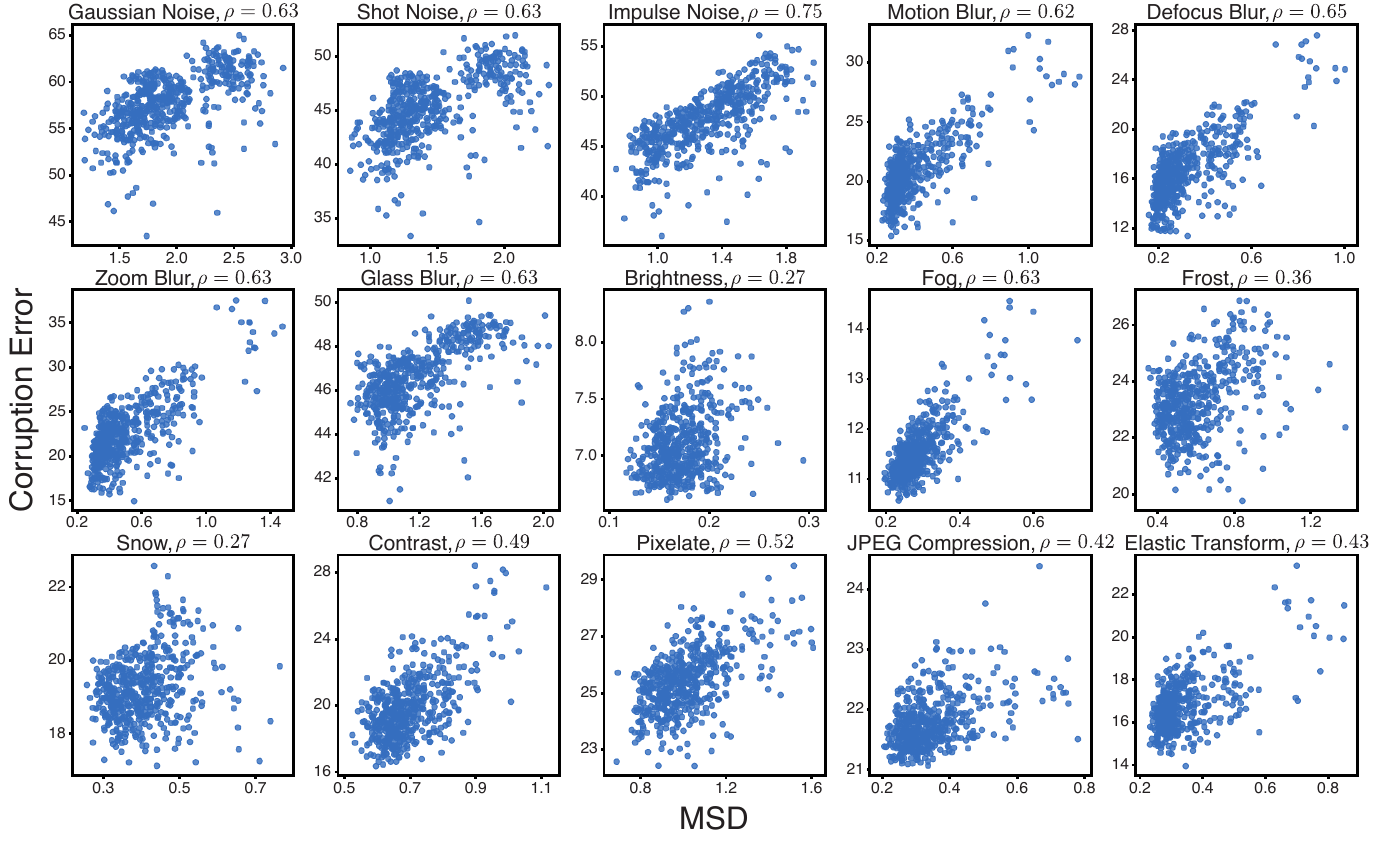}
    \caption{MSD vs. test error on the specified corruption. The test error is obtained by adapting a model's batch norm statistics to the augmented data distribution. The Spearman rank coefficient is given in parenthesis for each corruption. Correlation is still strong but weaker than training the models from scratch on the augmentations.}
    \label{fig:bnadapt}
\end{figure}

It is suggested in \citet{schneider2020improving} that significant improvement on a set of corruptions may be obtained by adapting only the batch norm parameters of a model trained on clean data to the statistics of the corrupted dataset. One might then expect that there will be a correlation between augmentation-corruption MSD and the error of a model whose batch norm has been adapted to the augmentation distribution. Such a correlation would suggest that a significant benefit of performing augmentations comes from making the batch norm statistics of the training set more similar to the corruption set. Here we test this, performing batch norm adaptation as described in \citet{schneider2020improving}, starting from a model trained with default CIFAR-10 augmentation. We choose the one hyperparameter in their algorithm such that the batch norm parameters are adapted completely to the augmented data distribution. Results are shown in Figure \ref{fig:bnadapt}. We find that this still correlates well with MSD (though, as is to be expected, less well than training on the augmentations). This lends support to the claim that batch norm statistics are an important aspect of the choice of augmentation.
\\
\section{MSD Ablation}
\label{app:ablation}
\subsection{Architecture choice}
\begin{figure}
    \centering
    \includegraphics[scale=1.0]{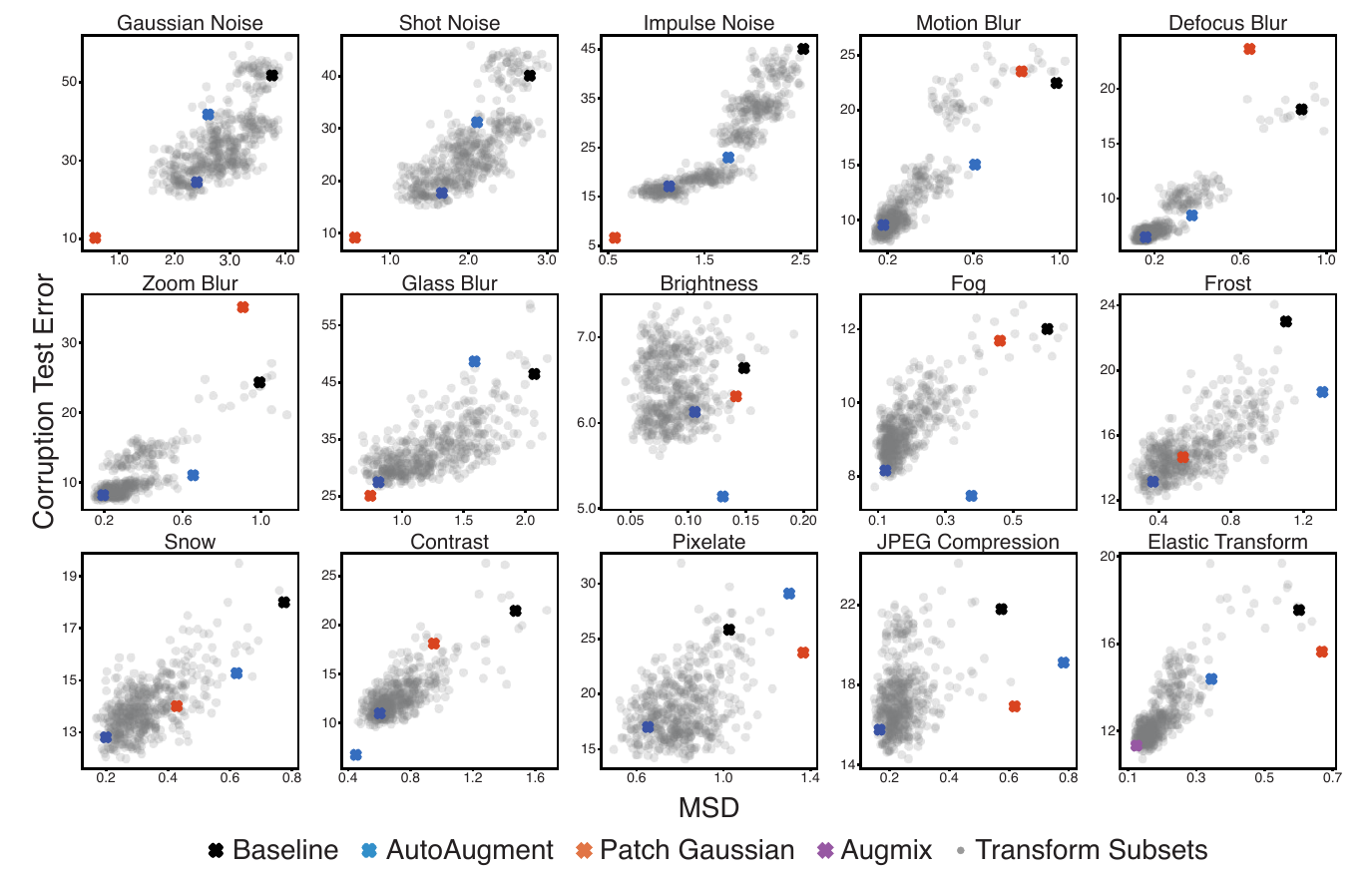}
    \caption{MSD vs corruption test error for which MSD is calculated using VGG-19-BN as the architecture for feature extraction.  The corruption error is still calculated using WideResNet-40-2.  Compare to Figure \ref{fig:newdsetshifts} to see that the qualitative structure of the correlation is the regardless of which architecture is used for the feature extractor.}
    \label{fig:msdvgg}
\end{figure}

Here we provide evidence that changing the architecture of the feature extractor used in the definition of MSD does not have any qualitative effect on the correlation with corruption error.  We use a version of VGG-19 with batch normalization that has been modified for CIFAR-10.  Otherwise, all other parameters are chosen the same.  We then repeat the experiment of Section \ref{sec:simpredictserr}.  In Table \ref{tab:spearmanrhovgg} and Figure \ref{fig:msdvgg}, we show that the qualitative results of this experiment are unchanged when using VGG-19-BN as the feature extractor.
\begin{table}
    \caption{Spearman's rank coefficient for the correlation between MSD and corruption error for two architectures in the feature extractor: WideResNet-40-2 and VGG-19-BN.  While WideResNet has slightly better correlations overall, the relative behavior across corruptions remains the same for the two architectures.}
    \label{tab:spearmanrhovgg}
    \centering
    \small
    \begin{tabular}[t]{z{70}  c  c}
      Corruption & WRN & VGG \\ \shline
      Gaussian Noise & 0.76 & 0.70\\
      Shot Noise & 0.83 & 0.78\\
      Impulse Noise & 0.90 & 0.92 \\
      Motion Blur & 0.86 & 0.81\\
      Defocus Blur & 0.83 & 0.78 \\
      Zoom Noise & 0.77& 0.68 \\
      Glass Blur & 0.69 & 0.66 \\
      Brightness & 0.27 & 0.08\\
   \end{tabular}
   \begin{tabular}[t]{z{70}  c  c}
      Corruption & WRN & VGG \\ \shline
      Fog & 0.68 & 0.60 \\
      Frost & 0.66 & 0.66\\
      Snow & 0.65 & 0.53\\
      Contrast & 0.66 & 0.65\\
      Pixelate & 0.35 & 0.29\\
      JPEG Compression & 0.33 & 0.26\\
      Elastic Transform & 0.77 & 0.74 \\
   \end{tabular}
\end{table}
\subsection{Parameter dependencies}
In calculating the feature space for transforms and MSD, it is necessary to both pick a number of images to average over and a number of corruptions to average over.  In our experiments, we use 100 images and 100 corruptions.  Here we provide evidence that these are reasonable choices for these parameters.

\begin{figure}
    \centering
    \includegraphics[scale=1.0]{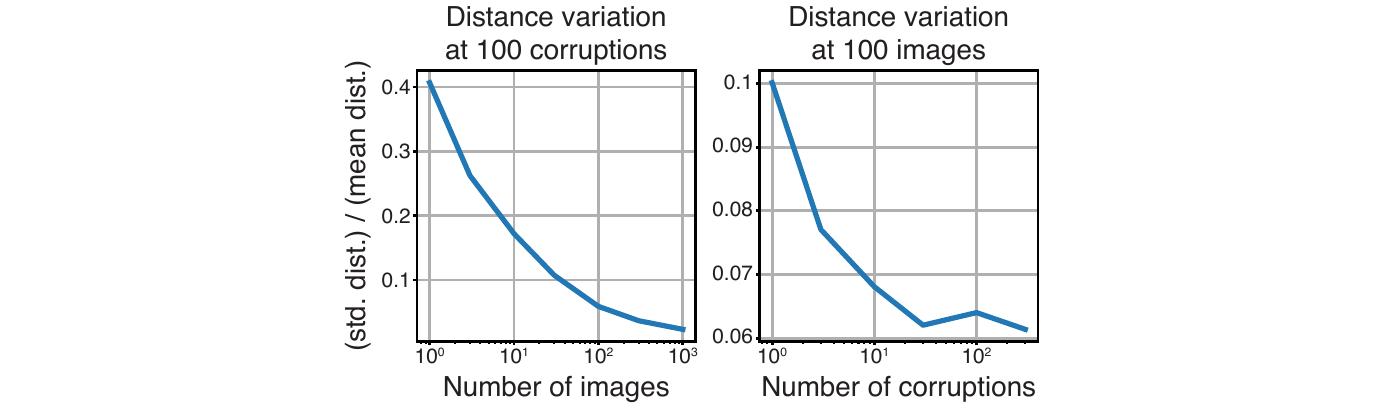}
    \caption{The standard deviation of the distance between an augmentation and a corruption center, taken over 100 resamplings of images and corruptions.  The standard deviation is calculated as a percentage of the mean distance, then averaged over 100 augmentation-corruption pairs.  At our choice of parameters, 100 images and 100 corruptions, the standard deviation is only  around 5\% of the distance.  This is smaller than the feature size in the scatter plots of Figure \ref{fig:distverror}}
    \label{fig:ablation}
\end{figure}

To do so, we use the augmentation scheme from AugMix and corruptions distributions from CIFAR-10-C to randomly sample 100 augmentation-corruption pairs.  Then, for different samplings of a fixed number of images and sampled corruptions, we measure the augmentation-corruption distance in the transform feature space 100 times for each augmentation-corruption pair.  We calculate the standard deviation of the distance as a percentage of the mean distance for each augmentation-corruption pair, and average this over pairs.  The results are shown in Figure \ref{fig:ablation}.  For our choice of image and corruption number, the standard deviation in distance is only around 5\% of the mean distance, which is smaller than the size of the features in the scatter plots in Figure \ref{fig:distverror}.

\section{ImageNet-\newC{} details}
\label{app:cbardetails}
\subsection{Dataset construction details}
\label{app:datasetconstruction}
First, 30 new corruptions, examples of which are shown in Figure \ref{fig:newcorruptions}, are adapted from common image filters and noise distributions available online \citep{jhlabs, filterpedia}.  These corruptions are generated in 10 severities such that the image remains human interpretable at all severities and the distribution of errors on a baseline model roughly matches that of ImageNet-C.

For each corruption, groups of 5 severities are generated that roughly match the average spread in error across severities in ImageNet-C on a baseline model.  Seven of these groups are formed for each corruption, each with one of severity 3 through 8 as the center severity of the group of 5.

A candidate dataset is a set of 10 groups of severities, each from a different corruption whose average corruption error on a baseline model is within 1\% of ImageNet-C.  This is necessary so that a relative decrease in error of data augmented models is normalized against a fixed baseline.  Also, more distorted, harder transforms are likely farther away, so if this wasn't fixed maximizing distance would likely just pick the hardest transforms in the highest severities.  It was computationally infeasible to enumerate all candidate datasets, so they were sampled as follows.  For each choice of 5 corruptions, one choice of severities was selected at random so that the average corruption error was within 1\% of ImageNet-C, if it existed.  Then random disjoint pairs of two sets of 5 were sampled to generate candidate datasets.  100k candidate datasets are sampled.

Call the set of all corruption-severity pairs in a dataset $\mathbb{C}$.  The distance of a candidate dataset to ImageNet-C is defined as
\begin{equation}
    d(\mathbb{C}_{\mathrm{new}}, \mathbb{C}_{\mathrm{IN-C}}) = \mathbb{E}_{c \sim \mathbb{C}_{\mathrm{new}}} \left [ \min_{c' \sim \mathbb{C}_{\mathrm{IN-C}}} d_{\mathrm{MMD}} (c, c') \right ] \, ,
\end{equation}
where $d_{\mathrm{MMD}}$ is defined in Section \ref{sec:transformsimilarity}.  The minimum helps assure that new corruptions are far from all ImageNet-C corruptions.

This distance is calculated for all 100k sampled candidate datasets.  For CIFAR-10, the same parameters described in Section \ref{sec:simpredictserr} are used to calculate the distance.  For ImageNet, the feature extractor is a ResNet-50 trained according to \citet{goyal2017accurate}, except color jittering is not used as a data augmentation.  Since there is much greater image diversity in ImageNet, we jointly sample 10k images and corruptions instead of independently sampling 100 images and 100 corruptions.  Code for measuring distances and training models is based on pyCls \citep{Radosavovic2019,Radosavovic2020}, and Hydra \citep{Yadan2019Hydra} is used for configuration.

\begin{figure}
    \centering
    \includegraphics[scale=1.0]{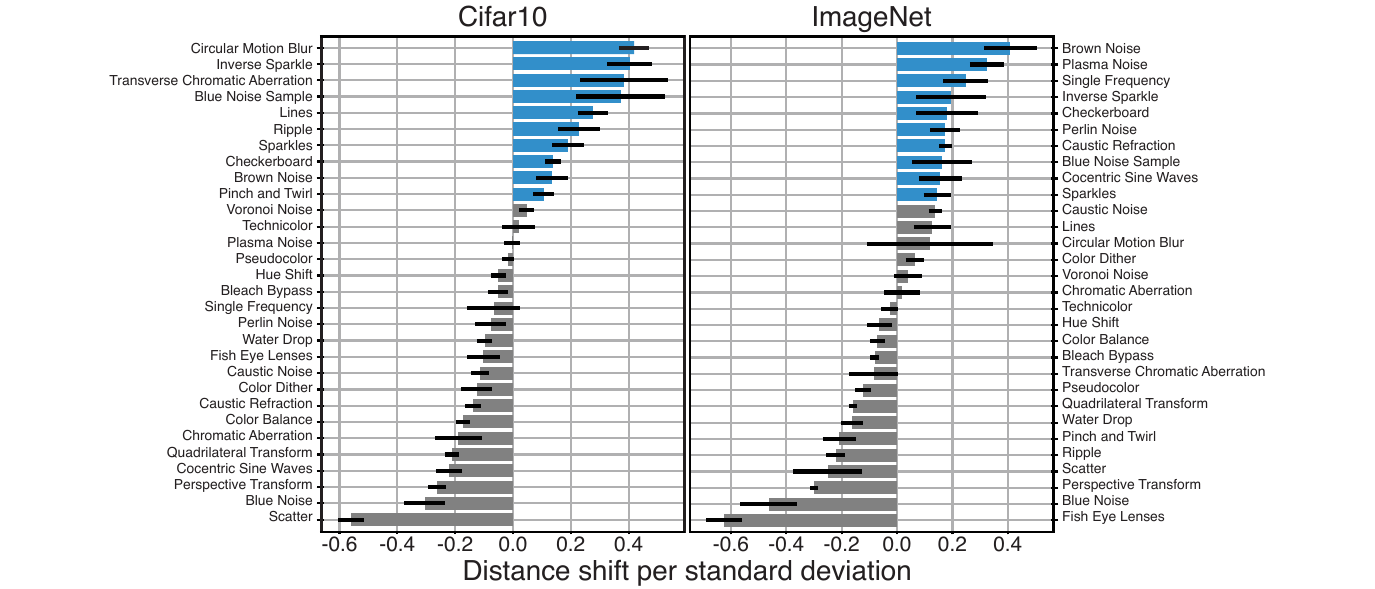}
    \caption{A corruption's average contribution to the distance to ImageNet-C, as a fraction of the population's standard deviation.  The blue corruptions are those used to construct ImageNet-\newC{}.}
    \label{fig:newdsetshifts}
\end{figure}

The corruptions are then ranked according the their average contribution to the dataset distance.  This entire procedure is repeated 10 times for CIFAR and 5 times for ImageNet, and corruption contributions are averaged.  The top 10 are chosen to form the new dataset.  These rankings are shown in Figure \ref{fig:newdsetshifts}. There may still be multiple candidate datasets made up of these 10 corruptions, differing by the choice of severities. Among these across all runs, we pick the one with error closest to ImageNet-C, though there may still be variation in error run-to-run.

\subsection{Complete results}
\label{app:completeresults}
Here we show results comparing ImageNet/CIFAR-10-C to ImageNet/CIFAR-10-\newC{}. The 10 transforms chosen for ImageNet-\newC{} are blue noise sample (BSmpl), plasma noise (Plsm), checkerboard (Ckbd), cocentric sine waves (CSin), single frequency (SFrq), brown noise (Brwn), perlin noise (Prln), inverse sparkle (ISprk), sparkles (Sprk), and caustic refraction (Rfrac).  For CIFAR-10-\newC{}, there is blue noise sample (BSmpl), brown noise (Brwn), checkerboard (Ckbd), circular motion blur (CBlur), inverse sparkle (ISprk), lines (Line), pinch and twirl (P\&T), ripple (Rppl), sparkles (Sprk), and transverse chromatic abberation (TCA). Table \ref{tab:appnewresultsmean} compares average results, representing the results from Table \ref{tab:newdatasetresults} in the main text for completeness. A breakdown of ImageNet/CIFAR-10-\newC{} results by corruption is in Table \ref{tab:appnewresultsall}, including standard deviations for each corruption individually. Stylized-ImageNet is trained jointly with ImageNet for half the epochs, as is done in \citet{geirhos2018imagenet}. ImageNet results are averaged over five runs, and CIFAR-10 over ten. For each of the five Stylized-ImageNet runs, we generate a new Stylized-ImageNet dataset using a different random seed and the code provided by \citet{geirhos2018imagenet}.

\subsection{MSD for CIFAR-10-\newC{}}
\begin{figure}
    \centering
    \includegraphics[scale=1.0]{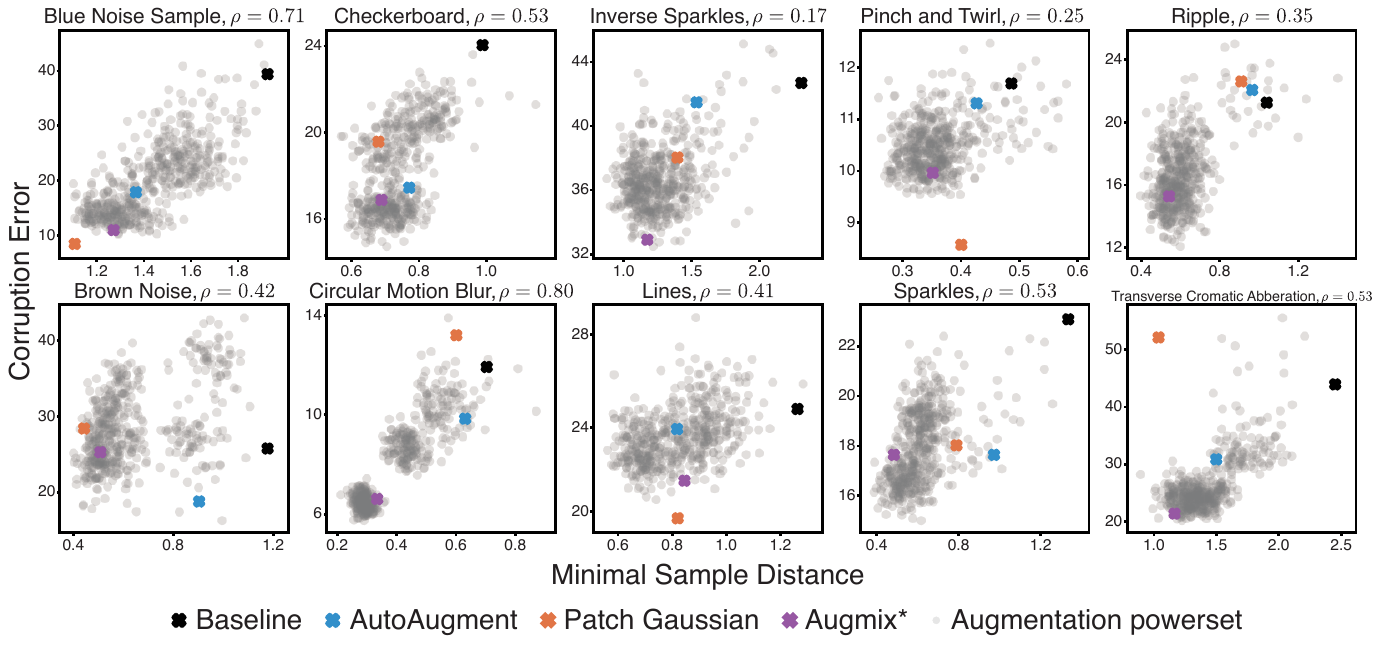}
    \caption{The correlation between MSD and corruption error on the dataset CIFAR-10-\newC{}. $\rho$ is the Spearman rank correlation.}
    \label{fig:msdcifar10cbar}
\end{figure}

We repeat the experiment of Section \ref{sec:simpredictserr} of the main text that measures the correlation between MSD and corruption error using the new corruptions in CIFAR-10-\newC{}. Results are shown in Figure \ref{fig:msdcifar10cbar}. We find that the correlation is still quite strong for many corruptions, though, like in CIFAR-10-C, there are some corruptions such as \emph{inverse sparkles} where the correlation is weak.

\begin{table}
    \caption{Comparison between performance on ImageNet/CIFAR10-C and ImageNet/CIFAR10-\newC{}.  Standard deviations are over 10 runs for CIFAR-10 and 5 runs for ImageNet. *ANT, DeepAugment (DA), and DeepAugment+AugMix (DA+AM) results use the pre-trained models provided with the respective papers and thus have different training parameters and only one run.}
    \label{tab:appnewresultsmean}
    \centering
    \begin{subtable}{0.49\textwidth}
        \centering
    \begin{tabular}[t]{ x{36}  x{30}  x{30} g{36}}
        & \textbf{IN-C} & \multicolumn{2}{c }{\textbf{IN-\newC{}}} \\ 
        Aug & Err & Err & $\Delta$IN-C \\
        \shline
        Baseline &  \err{58.1}{0.4} & \err{57.7}{0.2} & -0.5  \\ \hline
        AA & \err{55.0}{0.2} & \err{55.7}{0.3} & +0.7  \\ 
        SIN & \err{52.4}{0.1} & \err{55.8}{0.3} & +3.4 \\ 
        AugMix & \err{49.2}{0.7} & \err{52.4}{0.2} & +3.2 \\ 
        PG & \err{49.3}{0.2} & \err{56.6}{0.4} & +7.3  \\ 
        ANT* & \phantomerr{48.8}{0.2} & \phantomerr{53.9}{0.2} &  +5.1  \\ 
        DA* & \phantomerr{46.6}{0.2} & \phantomerr{51.0}{0.2} &  +4.4  \\
        DA+AM* & \phantomerr{41.0}{0.2} & \phantomerr{48.3}{0.2} &  +7.3 \\
        
    \end{tabular}
    \end{subtable}
    \begin{subtable}{0.49\textwidth}
        \centering
    \begin{tabular}[t]{ x{36}  x{30}  x{30} g{36} }
        & \textbf{C10-C} & \multicolumn{2}{c }{\textbf{C10-\newC{}}}   \\ 
        Aug & Err & Err & $\Delta$C10-C  \\
        \shline
        Baseline & \err{27.0}{0.6} & \err{27.1}{0.5} & +0.1  \\ \hline
        AA & \err{19.4}{0.2} & \err{21.0}{0.4} & +1.6    \\ 
        AugMix & \err{11.1}{0.2} & \err{16.0}{0.3} & +4.9 \\
        PG &\err{17.0}{0.4} & \err{23.8}{0.5} & +6.8  \\ 
    \end{tabular}
    \end{subtable}
\end{table}

\begin{table}
    \caption{Breakdown of performance on individual corruptions in ImageNet/CIFAR10-\newC{}.  Standard deviations are over 10 runs for CIFAR-10 and 5 runs for ImageNet.  Examples and full names of each corruption are given in Appendix \ref{app:transforms}.  *ANT, DeepAugment (DA), and DeepAugment+AugMix (DA+AM) results use the pre-trained models provided with the respective papers and thus have different training parameters and only one run.}
    \label{tab:appnewresultsall}
    \centering
    \small
    \setlength\tabcolsep{1.5pt}
    \begin{tabular}[t]{ x{36}   x{30} x{30} x{30} x{30} x{30} x{30} x{30} x{30} x{30} x{30} }
        & \multicolumn{10}{c }{\textbf{ ImageNet-\newC{} Corruptions} }  \\ 
        Aug  & BSmpl & Plsm & Ckbd & CSin & SFrq & Brwn & Prln & ISprk & Sprk & Rfrac \\
        \shline
        Baseline &   \err{68.6}{0.5} & \err{71.7}{0.7} & \err{49.4}{0.6} & \err{84.7}{0.5} & \err{79.0}{0.8} & \err{37.5}{0.5} & \err{34.3}{0.1} & \err{32.4}{0.5} & \err{76.7}{0.2} & \err{42.8}{0.2} \\ \hline
        AA &  \err{54.8}{0.7} & \err{68.3}{0.7} & \err{43.8}{1.0} & \err{86.5}{0.6} & \err{78.8}{0.9} & \err{34.5}{0.8} & \err{33.8}{0.2} & \err{36.1}{1.0} & \err{77.1}{1.2} & \err{43.8}{0.2} \\ 
        SIN & \err{54.7}{1.5} & \err{69.8}{1.1} & \err{52.8}{1.0} & \err{79.6}{0.4} & \err{69.2}{0.6} & \err{37.8}{0.4} & \err{35.3}{0.1} & \err{37.0}{0.5} & \err{77.3}{0.8} & \err{44.1}{0.2} \\ 
        AugMix & \err{43.2}{0.8} & \err{72.2}{0.4} & \err{46.1}{0.2} & \err{76.3}{0.3} & \err{67.4}{0.7} & \err{38.8}{0.5} & \err{32.4}{0.1} & \err{32.3}{0.2} & \err{76.4}{0.4} & \err{39.2}{0.2} \\ 
        PG  & \err{60.3}{2.9} & \err{74.1}{0.7} & \err{48.5}{1.0} & \err{82.1}{0.4} & \err{76.7}{0.8} & \err{38.9}{0.4} & \err{34.6}{0.1} & \err{32.1}{0.7} & \err{76.5}{0.6} & \err{42.1}{0.4} \\ 
        ANT* &  \phantomerr{35.8}{0.2} & \phantomerr{75.5}{0.2} & \phantomerr{56.9}{0.2} & \phantomerr{76.4}{0.2} & \phantomerr{63.7}{0.2} & \phantomerr{41.0}{0.2} & \phantomerr{35.2}{0.2} & \phantomerr{35.0}{0.2} & \phantomerr{76.1}{0.2} & \phantomerr{43.3}{0.2} \\ 
                DA*  & \phantomerr{41.7}{0.2} & \phantomerr{73.3}{0.2} & \phantomerr{53.9}{0.2} & \phantomerr{74.6}{0.2} & \phantomerr{50.9}{0.2} & \phantomerr{37.2}{0.2} & \phantomerr{30.3}{0.2} & \phantomerr{32.9}{0.2} & \phantomerr{74.7}{0.2} & \phantomerr{40.9}{0.2} \\ 
         DA+AM*  & \phantomerr{34.9}{0.2} & \phantomerr{67.9}{0.2} & \phantomerr{49.8}{0.2} & \phantomerr{69.7}{0.2} & \phantomerr{48.0}{0.2} & \phantomerr{35.2}{0.2} & \phantomerr{30.6}{0.2} & \phantomerr{32.9}{0.2} & \phantomerr{74.3}{0.2} & \phantomerr{39.8}{0.2} \\ 
               
    \end{tabular}
    \\
    \begin{tabular}[t]{ x{36}   x{30} x{30} x{30} x{30} x{30} x{30} x{30} x{30} x{30} x{30} }
         & \multicolumn{10}{c }{\textbf{CIFAR-10-\newC{} Corruptions}}  \\ 
        Aug & BSmpl & Brwn & Ckbd & CBlur & ISprk & Line & P\&T & Rppl & Sprk & TCA \\
        \shline
        Baseline &  \err{42.9}{5.1} & \err{27.2}{0.5} & \err{23.3}{0.6} & \err{11.8}{0.4} & \err{43.3}{0.8} & \err{26.2}{0.9} & \err{11.3}{0.3} & \err{21.6}{1.2} & \err{21.0}{1.1} & \err{42.9}{2.7} \\ \hline
        AA & \err{17.7}{1.7} & \err{17.5}{0.5} & \err{17.6}{0.5} & \err{9.5}{0.3} & \err{40.4}{1.5} & \err{23.6}{0.7} & \err{10.7}{0.3} & \err{23.5}{0.5} & \err{17.5}{0.7} & \err{31.8}{1.8}  \\ 
        AugMix & \err{9.8}{0.7} & \err{27.8}{1.3} & \err{13.4}{0.4} & \err{5.9}{0.2} & \err{30.3}{0.7} & \err{18.0}{0.6} & \err{8.3}{0.2} & \err{12.1}{0.4} & \err{15.5}{0.5} & \err{19.2}{1.0} \\
        PG & \err{9.0}{1.1} & \err{30.1}{1.1} & \err{21.6}{0.8} & \err{12.8}{0.5} & \err{35.4}{1.6} & \err{20.6}{0.5} & \err{8.8}{0.2} & \err{21.5}{0.9} & \err{19.3}{0.5} & \err{59.5}{3.5} \\ 
    \end{tabular}
\end{table}

\section{Resource usage}
WideResNet-40-2 on CIFAR-10 is trained for about 45 minutes to an hour on 1 V100 GPU, while ResNet-50 on ImageNet is trained for approximately 20 hours on 8 V100 GPUS. Collecting augmentation features for MSD requires 45 to an hour on 1 V100 GPU. In-memory corruption evaluation and feature extraction for CIFAR-10/ImageNet-C and the newly introduced corruptions is often CPU limited and runtimes vary significantly from corruption type to corruption type. This ranges up to approximately 6 hours on 80 Intel Xenon 2.2Ghz CPUs for per corruption and severity for ImageNet, or up to approximately 8 minutes per corruption and severity on 40 CPUs for CIFAR-10. When calculating distances for choosing CIFAR-10/ImageNet-\newC{}, CIFAR-10 uses the same amount of time per corruption as evaluation of the corruption, while ImageNet uses 1/5th the time, simply as a result of the number of images processed in each case.

\section{Glossary of transforms}
\label{app:transforms}
\begin{figure}
    \centering
    \includegraphics[scale=1.0]{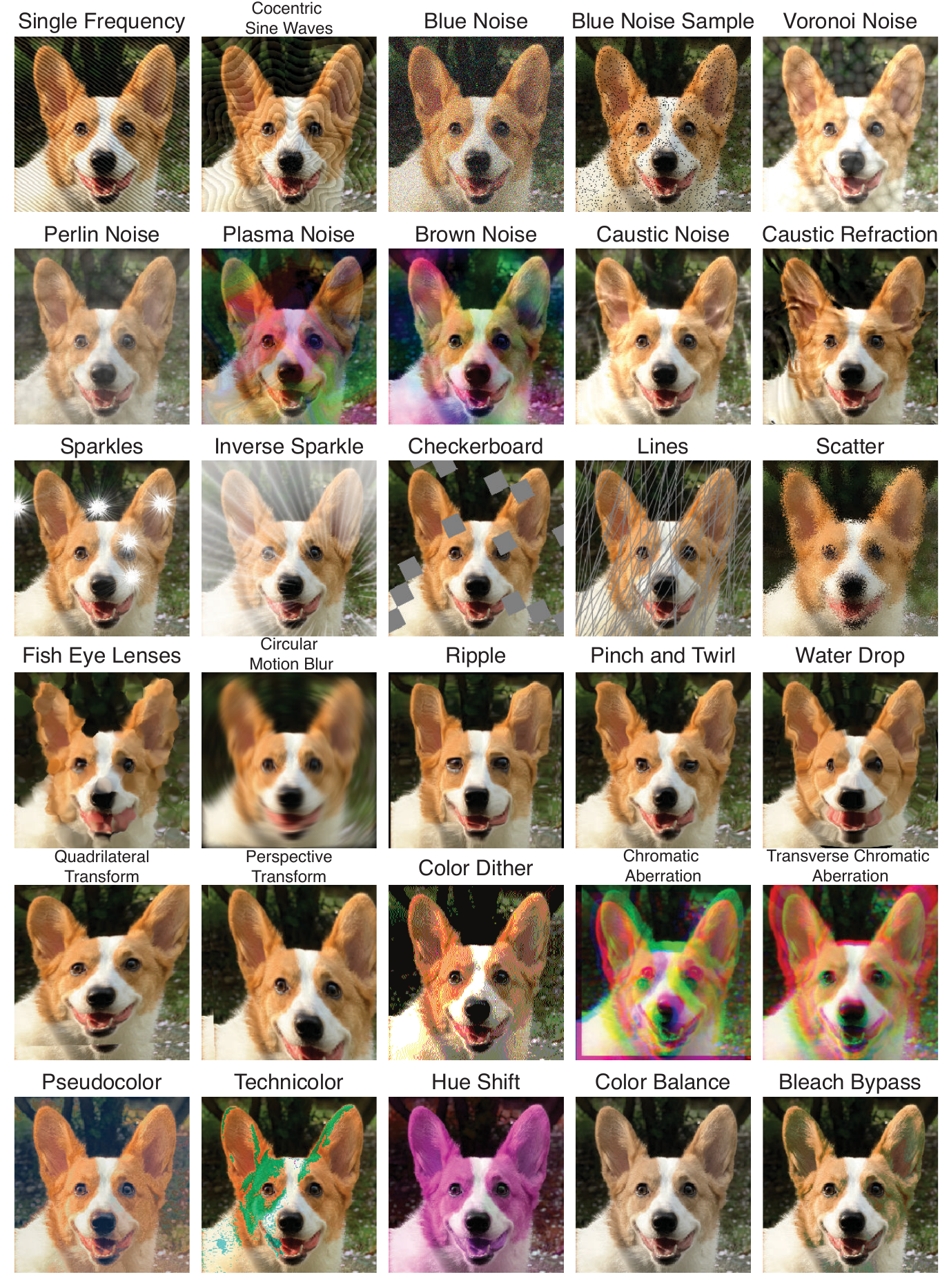}
    \caption{Examples of each corruption considered when building the dataset dissimilar to ImageNet-C. Base image \copyright~Sehee Park.}
    \label{fig:newcorruptions}
\end{figure}

This appendix contains examples of the augmentations and corruptions discussed in the text.  Figure \ref{fig:newcorruptions} shows the 30 new corruptions introduced in Section \ref{sec:newdataset}.  These transforms are adapted from common online filters and noise sources \citep{jhlabs,filterpedia}.  They are designed to be human interpretable and cover a wide range transforms, including noise additions, obscuring, warping, and color shifts.

\begin{figure}
    \centering
    \includegraphics[scale=1.0]{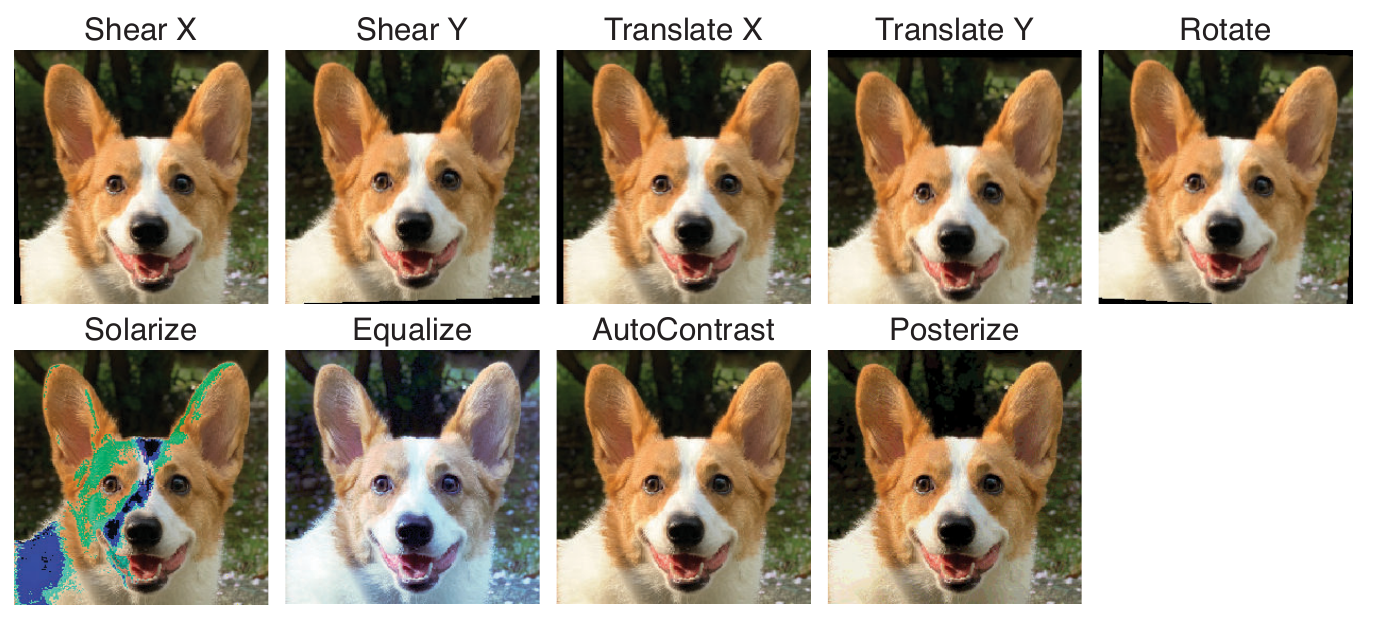}
    \caption{The nine base transforms used as augmentations in analysis. Base image \copyright~Sehee Park.}
    \label{fig:augmentations}
\end{figure}

Figure \ref{fig:augmentations} shows the 9 base transforms used to build augmentation schemes in the analysis.  These are transforms from the Pillow Image Library that are often used as data augmentation.  They have no exact overlap with either the corruptions of ImageNet-C or the new corruptions we introduce here.  There are five geometric transforms (shear x/y, translate x/y, and rotate) and four color transforms (solarize, equalize, autocontrast, and posterize).  We choose this particular set of augmentations following \citet{hendrycks2019augmix}.

\begin{figure}
    \centering
    \includegraphics[scale=1.0]{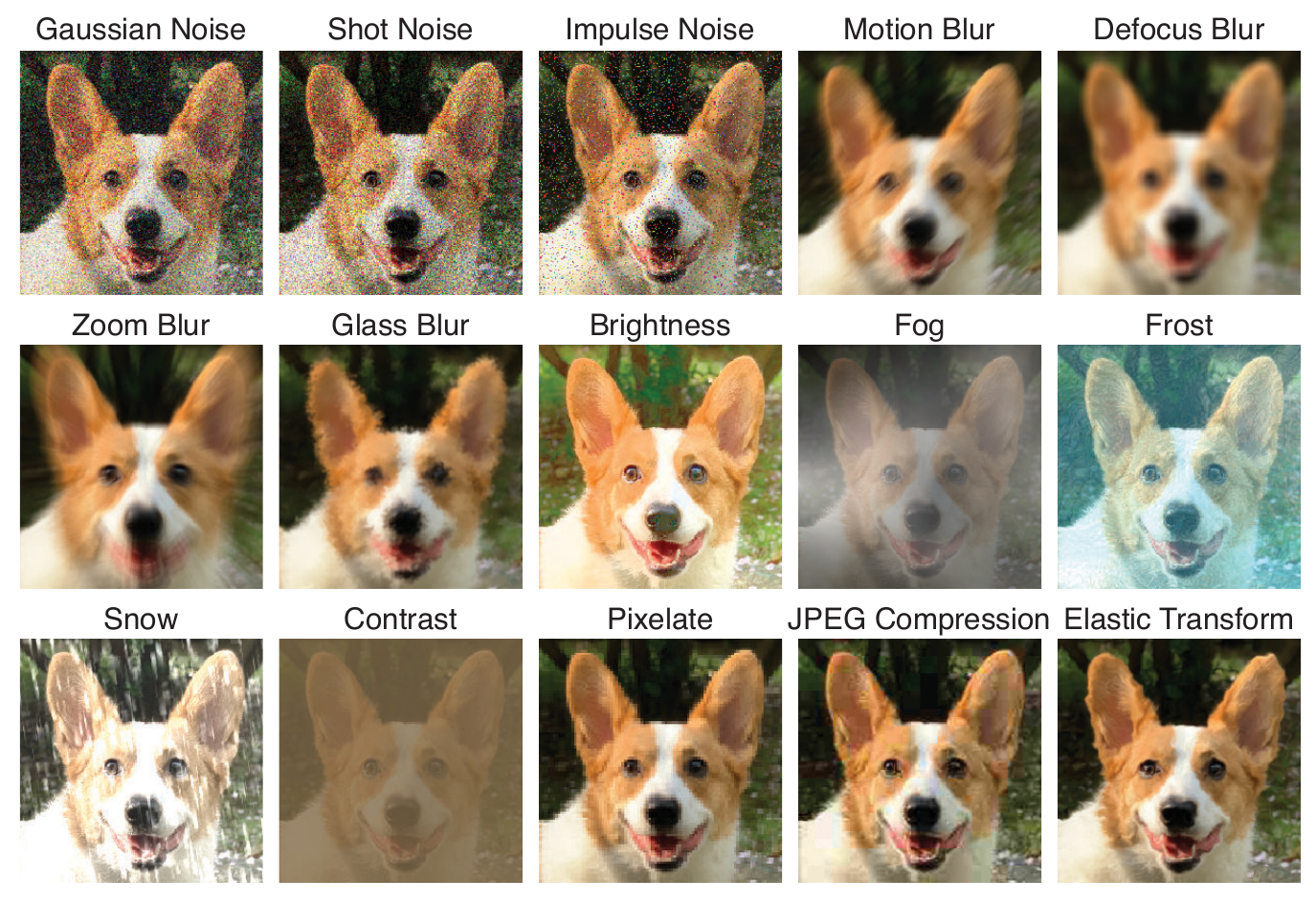}
    \caption{Examples of the 15 corruptions in the ImageNet-C corruption benchmark \citep{hendrycks2018benchmarking}. Base image \copyright~Sehee Park.}
    \label{fig:imagenetc}
\end{figure}

Figure \ref{fig:imagenetc} shows example corruptions from the ImageNet-C benchmark \citep{hendrycks2018benchmarking}.  They a grouped into four categories: noise (gaussian noise, shot noise, and impulse noise), blurs (motion blur, defocus blur, zoom blur, and glass blur), synthetic weather effects (brightness, fog, frost, and snow), and digital transforms (contrast, pixelate, JPEG compression, and elastic transform).

\end{document}

%% file: math_commands.tex
\usepackage{amsmath,amsfonts,bm}

\def\eqref#1{equation~\ref{#1}}

\def\1{\bm{1}}

\DeclareMathAlphabet{\mathsfit}{\encodingdefault}{\sfdefault}{m}{sl}
\SetMathAlphabet{\mathsfit}{bold}{\encodingdefault}{\sfdefault}{bx}{n}

\def\sA{{\mathbb{A}}}

\def\sD{{\mathbb{D}}}

\newcommand{\E}{\mathbb{E}}



%% file: robustness_neurips_arxiv.bbl
\begin{thebibliography}{46}
\providecommand{\natexlab}[1]{#1}
\providecommand{\url}[1]{\texttt{#1}}
\expandafter\ifx\csname urlstyle\endcsname\relax
  \providecommand{\doi}[1]{doi: #1}\else
  \providecommand{\doi}{doi: \begingroup \urlstyle{rm}\Url}\fi

\bibitem[Bruna \& Mallat(2013)Bruna and Mallat]{bruna2013invariant}
Bruna, J. and Mallat, S.
\newblock Invariant scattering convolution networks.
\newblock \emph{IEEE transactions on pattern analysis and machine
  intelligence}, 35\penalty0 (8):\penalty0 1872--1886, 2013.

\bibitem[Cubuk et~al.(2019)Cubuk, Zoph, Man{\'e}, Vasudevan, and
  Le]{cubuk2019autoaugment}
Cubuk, E.~D., Zoph, B., Man{\'e}, D., Vasudevan, V., and Le, Q.~V.
\newblock {AutoAugment: Learning augmentation strategies from data}.
\newblock In \emph{CVPR}, 2019.

\bibitem[Dao et~al.(2019)Dao, Gu, Ratner, Smith, De~Sa, and
  R{\'e}]{dao2019kernel}
Dao, T., Gu, A., Ratner, A.~J., Smith, V., De~Sa, C., and R{\'e}, C.
\newblock {A kernel theory of modern data augmentation}.
\newblock \emph{Proceedings of machine learning research}, 97:\penalty0 1528,
  2019.

\bibitem[Deng et~al.(2009)Deng, Dong, Socher, Li, Li, and
  Fei-Fei]{deng2009imagenet}
Deng, J., Dong, W., Socher, R., Li, L.-J., Li, K., and Fei-Fei, L.
\newblock {ImageNet: A large-scale hierarchical image database}.
\newblock In \emph{CVPR}, 2009.

\bibitem[Ding et~al.(2020)Ding, Ma, Wang, and Simoncelli]{ding2020image}
Ding, K., Ma, K., Wang, S., and Simoncelli, E.~P.
\newblock {Image quality assessment: Unifying structure and texture
  similarity}.
\newblock \emph{IEEE transactions on pattern analysis and machine
  intelligence}, 2020.

\bibitem[Dodge \& Karam(2017)Dodge and Karam]{dodge2017study}
Dodge, S. and Karam, L.
\newblock A study and comparison of human and deep learning recognition
  performance under visual distortions.
\newblock In \emph{ICCCN}, 2017.

\bibitem[Dosovitskiy et~al.(2021)Dosovitskiy, Beyer, Kolesnikov, Weissenborn,
  Zhai, Unterthiner, Dehghani, Minderer, Heigold, Gelly, Uszkoreit, and
  Houlsby]{dosovitskiy2020image}
Dosovitskiy, A., Beyer, L., Kolesnikov, A., Weissenborn, D., Zhai, X.,
  Unterthiner, T., Dehghani, M., Minderer, M., Heigold, G., Gelly, S.,
  Uszkoreit, J., and Houlsby, N.
\newblock {An image is worth 16x16 words: Transformers for image recognition at
  scale}.
\newblock In \emph{ICLR}, 2021.

\bibitem[Geirhos et~al.(2018)Geirhos, Temme, Rauber, Sch{\"u}tt, Bethge, and
  Wichmann]{geirhos2018generalisation}
Geirhos, R., Temme, C.~R., Rauber, J., Sch{\"u}tt, H.~H., Bethge, M., and
  Wichmann, F.~A.
\newblock Generalisation in humans and deep neural networks.
\newblock In \emph{NeurIPS}, 2018.

\bibitem[Geirhos et~al.(2019)Geirhos, Rubisch, Michaelis, Bethge, Wichmann, and
  Brendel]{geirhos2018imagenet}
Geirhos, R., Rubisch, P., Michaelis, C., Bethge, M., Wichmann, F.~A., and
  Brendel, W.
\newblock {ImageNet-trained CNNs are biased towards texture; increasing shape
  bias improves accuracy and robustness}.
\newblock In \emph{ICLR}, 2019.

\bibitem[Gladman(2016)]{filterpedia}
Gladman, S.~J.
\newblock Filterpedia, 2016.
\newblock URL \url{https://github.com/FlexMonkey/Filterpedia}.

\bibitem[Goyal et~al.(2017)Goyal, Doll{\'a}r, Girshick, Noordhuis, Wesolowski,
  Kyrola, Tulloch, Jia, and He]{goyal2017accurate}
Goyal, P., Doll{\'a}r, P., Girshick, R., Noordhuis, P., Wesolowski, L., Kyrola,
  A., Tulloch, A., Jia, Y., and He, K.
\newblock {Accurate, large minibatch SGD: Training ImageNet in 1 hour}.
\newblock \emph{arXiv preprint arXiv:1706.02677}, 2017.

\bibitem[He et~al.(2016)He, Zhang, Ren, and Sun]{he2016deep}
He, K., Zhang, X., Ren, S., and Sun, J.
\newblock Deep residual learning for image recognition.
\newblock In \emph{CVPR}, 2016.

\bibitem[Hendrycks \& Dietterich(2018)Hendrycks and
  Dietterich]{hendrycks2018benchmarking}
Hendrycks, D. and Dietterich, T.
\newblock Benchmarking neural network robustness to common corruptions and
  perturbations.
\newblock In \emph{ICLR}, 2018.

\bibitem[Hendrycks et~al.(2019)Hendrycks, Mu, Cubuk, Zoph, Gilmer, and
  Lakshminarayanan]{hendrycks2019augmix}
Hendrycks, D., Mu, N., Cubuk, E.~D., Zoph, B., Gilmer, J., and
  Lakshminarayanan, B.
\newblock {AugMix: A simple data processing method to improve robustness and
  uncertainty}.
\newblock In \emph{ICLR}, 2019.

\bibitem[Hendrycks et~al.(2020)Hendrycks, Basart, Mu, Kadavath, Wang, Dorundo,
  Desai, Zhu, Parajuli, Guo, Song, Steinhardt, and Gilmer]{hendrycks2020many}
Hendrycks, D., Basart, S., Mu, N., Kadavath, S., Wang, F., Dorundo, E., Desai,
  R., Zhu, T., Parajuli, S., Guo, M., Song, D., Steinhardt, J., and Gilmer, J.
\newblock {The many faces of robustness: A critical analysis of
  out-of-distribution generalization}.
\newblock \emph{arXiv preprint arXiv:2006.16241}, 2020.

\bibitem[Huxtable(2006)]{jhlabs}
Huxtable, J.
\newblock {JH Labs Java Image Processing}, 2006.
\newblock URL \url{http://www.jhlabs.com/ip/filters/}.

\bibitem[Krizhevsky et~al.(2009)Krizhevsky, Hinton,
  et~al.]{krizhevsky2009learning}
Krizhevsky, A., Hinton, G., et~al.
\newblock Learning multiple layers of features from tiny images.
\newblock 2009.

\bibitem[Lee et~al.(2020)Lee, Won, and Hong]{lee2020compounding}
Lee, J., Won, T., and Hong, K.
\newblock Compounding the performance improvements of assembled techniques in a
  convolutional neural network.
\newblock \emph{arXiv preprint arXiv:2001.06268}, 2020.

\bibitem[Liang et~al.(2020)Liang, Hu, and Feng]{liang2020we}
Liang, J., Hu, D., and Feng, J.
\newblock Do we really need to access the source data? source hypothesis
  transfer for unsupervised domain adaptation.
\newblock In \emph{International Conference on Machine Learning}, pp.\
  6028--6039. PMLR, 2020.

\bibitem[Lopes et~al.(2019)Lopes, Yin, Poole, Gilmer, and
  Cubuk]{lopes2019improving}
Lopes, R.~G., Yin, D., Poole, B., Gilmer, J., and Cubuk, E.~D.
\newblock {Improving robustness without sacrificing accuracy with Patch
  Gaussian augmentation}.
\newblock \emph{arXiv preprint arXiv:1906.02611}, 2019.

\bibitem[Mahajan et~al.(2018)Mahajan, Girshick, Ramanathan, He, Paluri, Li,
  Bharambe, and van~der Maaten]{mahajan2018exploring}
Mahajan, D., Girshick, R., Ramanathan, V., He, K., Paluri, M., Li, Y.,
  Bharambe, A., and van~der Maaten, L.
\newblock Exploring the limits of weakly supervised pretraining.
\newblock In \emph{Proceedings of the European Conference on Computer Vision
  (ECCV)}, pp.\  181--196, 2018.

\bibitem[Orhan(2019)]{orhan2019robustness}
Orhan, A.~E.
\newblock {Robustness properties of Facebook's ResNeXt WSL models}.
\newblock \emph{arXiv preprint arXiv:1907.07640}, 2019.

\bibitem[Radosavovic et~al.(2019)Radosavovic, Johnson, Xie, Lo, and
  Doll{\'a}r]{Radosavovic2019}
Radosavovic, I., Johnson, J., Xie, S., Lo, W.-Y., and Doll{\'a}r, P.
\newblock On network design spaces for visual recognition.
\newblock In \emph{ICCV}, 2019.

\bibitem[Radosavovic et~al.(2020)Radosavovic, Kosaraju, Girshick, He, and
  Doll{\'a}r]{Radosavovic2020}
Radosavovic, I., Kosaraju, R.~P., Girshick, R., He, K., and Doll{\'a}r, P.
\newblock Designing network design spaces.
\newblock In \emph{CVPR}, 2020.

\bibitem[Recht et~al.(2018)Recht, Roelofs, Schmidt, and
  Shankar]{recht2018cifar}
Recht, B., Roelofs, R., Schmidt, L., and Shankar, V.
\newblock {Do CIFAR-10 classifiers generalize to CIFAR-10?}
\newblock \emph{arXiv preprint arXiv:1806.00451}, 2018.

\bibitem[Recht et~al.(2019)Recht, Roelofs, Schmidt, and
  Shankar]{recht2019imagenet}
Recht, B., Roelofs, R., Schmidt, L., and Shankar, V.
\newblock {Do ImageNet classifiers generalize to ImageNet?}
\newblock In \emph{ICML}, 2019.

\bibitem[Rusak et~al.(2020)Rusak, Schott, Zimmermann, Bitterwolf, Bringmann,
  Bethge, and Brendel]{rusak2020increasing}
Rusak, E., Schott, L., Zimmermann, R., Bitterwolf, J., Bringmann, O., Bethge,
  M., and Brendel, W.
\newblock A simple way to make neural networks robust against diverse image
  corruptions.
\newblock \emph{arXiv preprint arXiv:2001.06057}, 2020.

\bibitem[Schneider et~al.(2020)Schneider, Rusak, Eck, Bringmann, Brendel, and
  Bethge]{schneider2020improving}
Schneider, S., Rusak, E., Eck, L., Bringmann, O., Brendel, W., and Bethge, M.
\newblock Improving robustness against common corruptions by covariate shift
  adaptation.
\newblock In \emph{NeurIPS}, 2020.

\bibitem[Shankar et~al.(2019)Shankar, Dave, Roelofs, Ramanan, Recht, and
  Schmidt]{shankar2019image}
Shankar, V., Dave, A., Roelofs, R., Ramanan, D., Recht, B., and Schmidt, L.
\newblock Do image classifiers generalize across time?
\newblock \emph{arXiv preprint arXiv:1906.02168}, 2019.

\bibitem[Simard et~al.(1998)Simard, LeCun, Denker, and
  Victorri]{simard1998transformation}
Simard, P.~Y., LeCun, Y.~A., Denker, J.~S., and Victorri, B.
\newblock Transformation invariance in pattern recognition—tangent distance
  and tangent propagation.
\newblock In \emph{Neural networks: tricks of the trade}, pp.\  239--274.
  Springer, 1998.

\bibitem[Simonyan \& Zisserman(2015)Simonyan and Zisserman]{simonyan2014very}
Simonyan, K. and Zisserman, A.
\newblock Very deep convolutional networks for large-scale image recognition.
\newblock In \emph{ICLR}, 2015.

\bibitem[Sun et~al.(2020)Sun, Wang, Liu, Miller, Efros, and Hardt]{sun2020test}
Sun, Y., Wang, X., Liu, Z., Miller, J., Efros, A., and Hardt, M.
\newblock Test-time training with self-supervision for generalization under
  distribution shifts.
\newblock In \emph{International Conference on Machine Learning}, pp.\
  9229--9248. PMLR, 2020.

\bibitem[Szegedy et~al.(2014)Szegedy, Zaremba, Sutskever, Bruna, Erhan,
  Goodfellow, and Fergus]{szegedy2013intriguing}
Szegedy, C., Zaremba, W., Sutskever, I., Bruna, J., Erhan, D., Goodfellow, I.,
  and Fergus, R.
\newblock Intriguing properties of neural networks.
\newblock In \emph{ICLR}, 2014.

\bibitem[Tan \& Le(2019)Tan and Le]{tan2019efficientnet}
Tan, M. and Le, Q.
\newblock {EfficientNet: Rethinking model scaling for convolutional neural
  networks}.
\newblock In \emph{ICML}, 2019.

\bibitem[Taori et~al.(2020)Taori, Dave, Shankar, Carlini, Recht, and
  Schmidt]{taori2020measuring}
Taori, R., Dave, A., Shankar, V., Carlini, N., Recht, B., and Schmidt, L.
\newblock Measuring robustness to natural distribution shifts in image
  classification.
\newblock In \emph{NeurIPS}, 2020.

\bibitem[Touvron et~al.(2020)Touvron, Cord, Douze, Massa, Sablayrolles, and
  J\'egou]{touvron2020deit}
Touvron, H., Cord, M., Douze, M., Massa, F., Sablayrolles, A., and J\'egou, H.
\newblock {Training data-efficient image transformers \& distillation through
  attention}.
\newblock \emph{arXiv preprint arXiv:2012.12877}, 2020.

\bibitem[Wang et~al.(2021)Wang, Shelhamer, Liu, Olshausen, and
  Darrell]{wang2020tent}
Wang, D., Shelhamer, E., Liu, S., Olshausen, B., and Darrell, T.
\newblock Tent: Fully test-time adaptation by entropy minimization.
\newblock \emph{ICLR}, 2021.

\bibitem[Wu et~al.(2020)Wu, Zhang, Valiant, and R{\'e}]{wu2020generalization}
Wu, S., Zhang, H.~R., Valiant, G., and R{\'e}, C.
\newblock On the generalization effects of linear transformations in data
  augmentation.
\newblock In \emph{ICML}, 2020.

\bibitem[Xie et~al.(2020)Xie, Luong, Hovy, and Le]{xie2020self}
Xie, Q., Luong, M.-T., Hovy, E., and Le, Q.~V.
\newblock {Self-training with Noisy Student improves imagenet classification}.
\newblock In \emph{CVPR}, 2020.

\bibitem[Yadan(2019)]{Yadan2019Hydra}
Yadan, O.
\newblock Hydra - a framework for elegantly configuring complex applications.
\newblock Github, 2019.
\newblock URL \url{https://github.com/facebookresearch/hydra}.

\bibitem[Yin et~al.(2019)Yin, Lopes, Shlens, Cubuk, and Gilmer]{yin2019fourier}
Yin, D., Lopes, R.~G., Shlens, J., Cubuk, E.~D., and Gilmer, J.
\newblock {A Fourier perspective on model robustness in computer vision}.
\newblock In \emph{NeurIPS}, 2019.

\bibitem[Zagoruyko \& Komodakis(2016)Zagoruyko and Komodakis]{BMVC2016_87}
Zagoruyko, S. and Komodakis, N.
\newblock Wide residual networks.
\newblock In \emph{BMVC}, 2016.

\bibitem[Zhang et~al.(2018{\natexlab{a}})Zhang, Cisse, Dauphin, and
  Lopez-Paz]{zhang2018mixup}
Zhang, H., Cisse, M., Dauphin, Y.~N., and Lopez-Paz, D.
\newblock mixup: Beyond empirical risk minimization.
\newblock In \emph{ICLR}, 2018{\natexlab{a}}.

\bibitem[Zhang et~al.(2020)Zhang, Wu, Zhang, Zhu, Zhang, Lin, Sun, He, Muller,
  Manmatha, Li, and Smola]{zhang2020resnest}
Zhang, H., Wu, C., Zhang, Z., Zhu, Y., Zhang, Z., Lin, H., Sun, Y., He, T.,
  Muller, J., Manmatha, R., Li, M., and Smola, A.
\newblock {ResNeSt: Split-Attention Networks}.
\newblock \emph{arXiv preprint arXiv:2004.08955}, 2020.

\bibitem[Zhang et~al.(2018{\natexlab{b}})Zhang, Isola, Efros, Shechtman, and
  Wang]{zhang2018unreasonable}
Zhang, R., Isola, P., Efros, A.~A., Shechtman, E., and Wang, O.
\newblock The unreasonable effectiveness of deep features as a perceptual
  metric.
\newblock In \emph{CVPR}, 2018{\natexlab{b}}.

\bibitem[Zilly et~al.(2019)Zilly, Zilly, Richter, Wattenhofer, Censi, and
  Frazzoli]{zilly2019frechet}
Zilly, J., Zilly, H., Richter, O., Wattenhofer, R., Censi, A., and Frazzoli, E.
\newblock {The Frechet Distance of training and test distribution predicts the
  generalization gap}.
\newblock \emph{OpenReview preprint}, 2019.
\newblock URL \url{https://openreview.net/forum?id=SJgSflHKDr}.

\end{thebibliography}
